\documentclass{article}

    \PassOptionsToPackage{numbers, sort&compress}{natbib}

    \usepackage[final]{neurips_2023}

\usepackage[utf8]{inputenc} 
\usepackage[T1]{fontenc}    
\usepackage{url}            
\usepackage{booktabs}       
\usepackage{amsfonts}       
\usepackage{nicefrac}       
\usepackage{microtype}      

\usepackage{graphicx}
\graphicspath{{assets/}} 
\usepackage{amsmath}
\usepackage{amssymb}
\usepackage{booktabs}
\usepackage{enumitem}
\usepackage{algorithm}
\usepackage{algpseudocode}
\usepackage{bm}
\usepackage{multirow}
\usepackage{caption}
\usepackage{subcaption}
\usepackage[table]{xcolor}
\definecolor{lightyellow}{RGB}{225,255,224}
\definecolor{mycolor}{RGB}{208,244,222}
\definecolor{lightgreen}{RGB}{144,238,144}
\definecolor{lightblue}{RGB}{173,216,230}
\definecolor{lightred}{RGB}{255,182,193}
\definecolor{lightpurple}{RGB}{147,112,219}
\definecolor{lightorange}{RGB}{255,165,0}
\usepackage{colortbl}
\definecolor{Gray}{gray}{0.85}

\usepackage{changepage} 

\usepackage[pagebackref,breaklinks,colorlinks]{hyperref}

\usepackage[capitalize]{cleveref}
\crefname{section}{Sec.}{Secs.}
\Crefname{section}{Section}{Sections}
\Crefname{table}{Table}{Tables}
\crefname{table}{Tab.}{Tabs.}

\newcommand{\bline}[1]{\textcolor{orange}{#1}}

\usepackage{xspace}
\makeatletter
\DeclareRobustCommand\onedot{\futurelet\@let@token\@onedot}
\def\@onedot{\ifx\@let@token.\else.\null\fi\xspace}

\def\eg{\emph{e.g}\onedot} 
\def\ie{\emph{i.e}\onedot}

\def\etal{\emph{et al}\onedot}
\makeatother

\title{Understand, Improve and Iterate: Towards Developing Explainable Metrics for Evaluating and Improving Fine-Grain Text-to-Image Alignment}
\title{Divide, Evaluate, and Conquer: Developing Explainable Metrics for Evaluating and Improving Fine-Grain Text-to-Image Alignment}
\title{Divide, Evaluate, and Refine: A Simple Decompositional Perspective on Evaluating and Improving Text-to-Image Alignment}
\title{Divide, Evaluate, and Refine:  Evaluating and Improving Text-to-Image Alignment with \\Iterative VQA Feedback}

\author{
Jaskirat Singh$^{1}$  \qquad \qquad \qquad \qquad \qquad  Liang Zheng$^{1,2}$\\
$^1$The Australian National University 
\qquad $^2$Australian Centre for Robotic Vision
\\
\url{https://1jsingh.github.io/divide-evaluate-and-refine}
}

\begin{document}

\onecolumn[
{
\vskip -0.2in
\maketitle
\begin{center}
\vskip -0.2in
\centerline{\includegraphics[width=1.\linewidth]{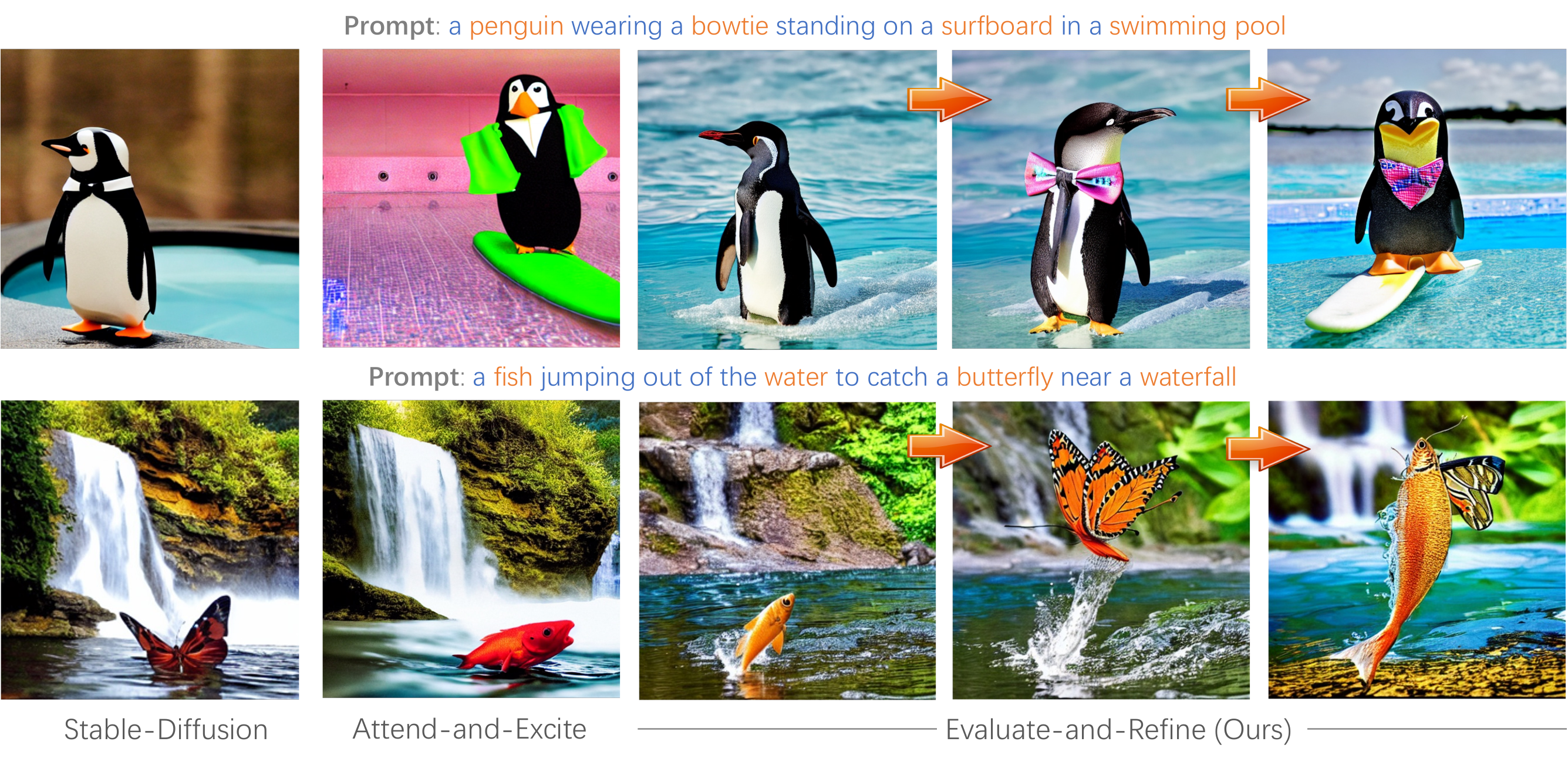}}
\vskip -0.05in
\captionsetup{type=figure}
    \caption{
    We propose a training-free decompositional framework which helps both better evaluate (Sec.~\ref{sec:vqa-score}) and gradually improve (Sec.~\ref{sec:vqa-iter}) text-to-image alignment using iterative VQA feedback.
    }
    \label{fig:overview}
\end{center}
}

\begin{abstract}

   The field of text-conditioned image generation has made unparalleled progress with the recent advent of latent diffusion models. While remarkable, as the complexity of given text input increases, the 
  state-of-the-art diffusion models may still fail in generating images which accurately convey the semantics of the given prompt. 
  Furthermore, it has been observed that such misalignments are often left undetected by pretrained multi-modal models such as \emph{CLIP}.
  To address these problems, in this paper we explore a
  simple yet effective
  decompositional approach towards both evaluation and improvement of text-to-image alignment. In particular,  we first introduce a
  \emph{\textbf{D}ecompositional-\textbf{A}lignment-\textbf{Score}} which given a complex prompt decomposes it into a set of disjoint assertions.  
  The alignment of each assertion with generated images is then measured using a VQA model. Finally, alignment scores for different assertions are combined aposteriori to give the final text-to-image alignment score. Experimental analysis reveals that the proposed alignment metric shows significantly higher correlation 
  with human ratings as opposed to traditional \emph{CLIP, BLIP} scores.
  Furthermore, we also find that the assertion level alignment scores provide a useful feedback which can then be used in a simple iterative procedure to gradually increase the expression of different assertions in the final image outputs. 
  Human user studies indicate that the proposed approach surpasses previous state-of-the-art by 8.7\% in overall text-to-image alignment accuracy.

\end{abstract}

\section{Introduction}
\label{sec:intro}

The field of text-to-image generation has made significant advancements with the recent advent of large-scale language-image (LLI) models
\cite{rombach2021highresolution,nichol2021glide,saharia2022photorealistic,ramesh2022hierarchical,yu2022scaling}. In particular, text-conditioned latent diffusion models have shown unparalleled success in generating creative imagery corresponding to a diverse range of free-form textual descriptions. However, while remarkable, it has been observed \cite{liu2022compositional,chefer2023attend,feng2022training} that as the complexity of the input text increases, the generated images do not always accurately align with the semantic meaning of the textual prompt. 





To facilitate the reliable use of current text-to-image generation models for practical applications,
it is essential to answer two key questions: 1) Can we detect such fine-grain misalignments between the input text and the generated output in a robust manner? and 2) Once detected, can we improve the text-to-image alignment for failure cases?
While several metrics for evaluating text-to-image alignment (\eg, CLIP \cite{hessel2021clipscore}, BLIP \cite{li2022blip}, BLIP2 \cite{li2023blip}) exist, 
it has been observed \cite{chefer2023attend,paiss2022no} that a high score with these metrics can be achieved even if the image does not fully correspond with input prompt. For instance, in Fig.~\ref{fig:overview}, an output image 
(containing only pink trees) 
shows high CLIP/BLIP scores with the text ``pink trees and yellow car'' even if yellow car is not present. Evaluating text-to-image matching using 
the 
image-text-matching (ITM) head of BLIP models has also been 
recently 
explored \cite{li2022blip,li2023blip}. However, the generated scores also show a similar tendency to favor the main subject of input prompt. Furthermore, even if such misalignments are detected, it is not clear how such information can be used for improving the quality of generated image outputs in a reliable manner.

\begin{figure}[t]
\vskip -0.1in
\begin{center}
\centering
     \begin{subfigure}[b]{0.495\columnwidth}
         \centering
         \includegraphics[width=1.\textwidth]{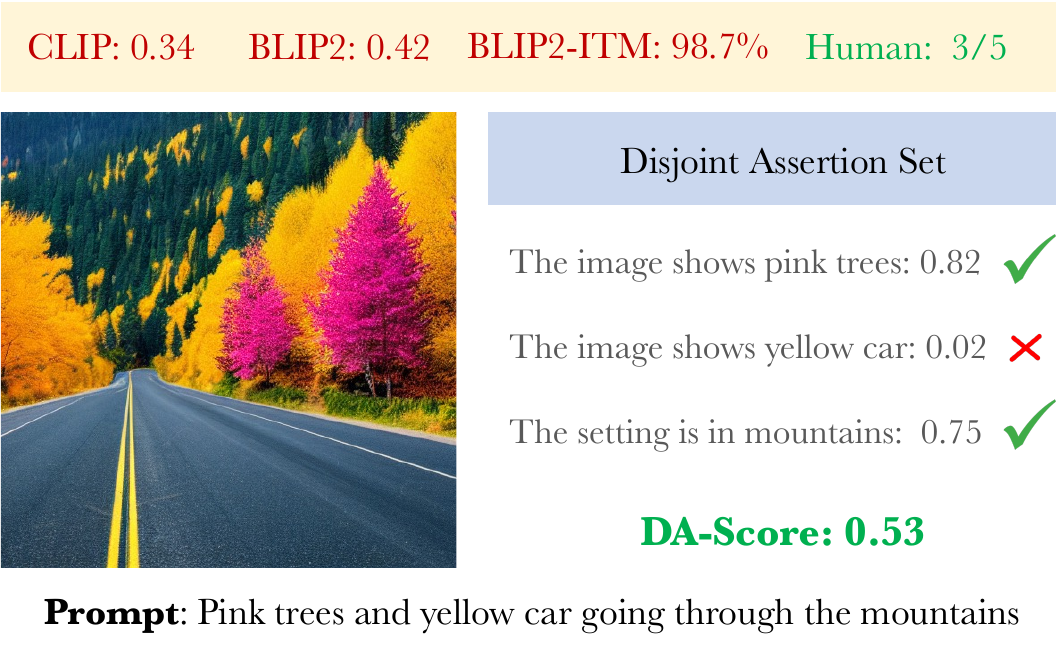}
     \end{subfigure}
     \hfill
      \begin{subfigure}[b]{0.495\columnwidth}
         \centering
         \includegraphics[width=1.\textwidth]{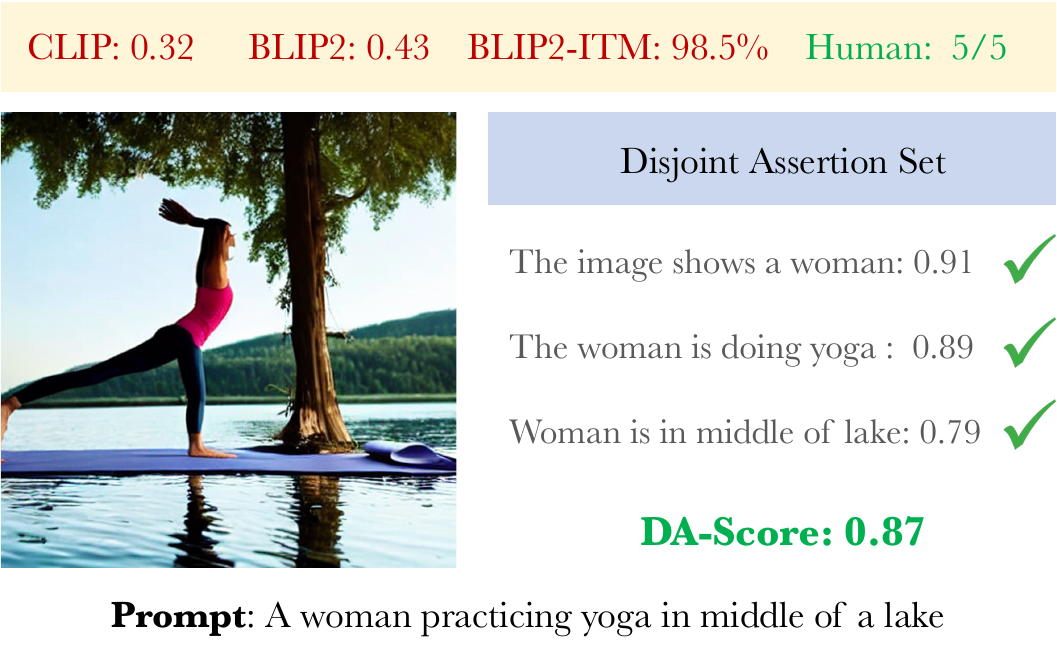}
     \end{subfigure} 
     \vskip 0.02in
     \begin{subfigure}[b]{1.0\columnwidth}
         \centering
         \includegraphics[width=1.\textwidth]{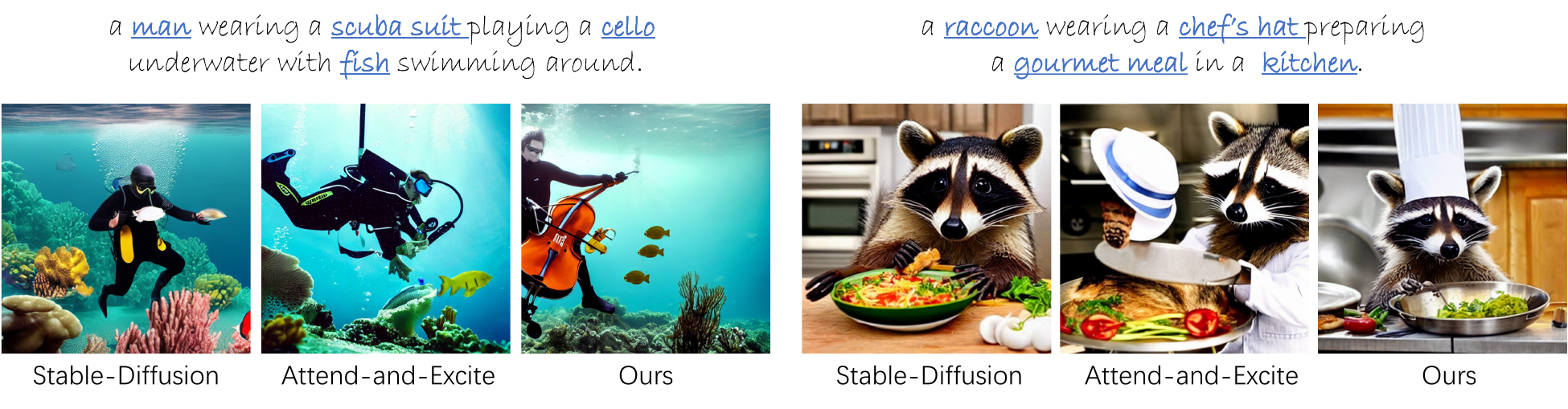}
     \end{subfigure} 
\vskip -0.05in
\caption{{\emph{Overview.}} \emph{Top:} Traditional methods for evaluating text-to-image alignment \eg, CLIP \cite{hessel2021clipscore}, BLIP-2 \cite{li2022blip} and BLIP2-ITM (which provides a binary image-text matching score between 0 and 1) 
often fail to distinguish between good (\emph{right}) and bad (\emph{left}) image outputs and
can give high scores even if  the generated image is not an accurate match for input prompt (missing yellow car). In contrast, by breaking down the prompt into a set of disjoint assertions and then evaluating their alignment with the generated image using a VQA model \cite{li2022blip}, the proposed Decompositional-Alignment Score (DA-score) shows much better correlation with human ratings (refer Sec.~\ref{sec:vqa-score-results}).  \emph{Bottom:} Furthermore, we show that the assertion-level alignment scores can be used along with a simple iterative refinement strategy to reliably improve the alignment of generated image outputs (refer Sec.~\ref{sec:vqa-iter-results}).}
\label{fig:overview}
\end{center}
\vskip -0.2in
\end{figure}

To address these problems, in this paper we explore a simple yet effective decompositional approach towards both evaluation and improvement of fine-grain text-to-image alignment. In particular, we propose a \emph{\textbf{D}ecompositional-\textbf{A}lignment-\textbf{Score}} (DA-Score) which 
given a complex text prompt, first decomposes it into a set of disjoint assertions about the content of the prompt. The alignment of each of these assertions with the generated image is then measured using a VQA model \cite{kim2021vilt,li2022blip}. Finally, the alignment scores for diffferent assertions are combined to give an overall text-to-image alignment score. 
Our experiments reveal that the proposed evaluation score shows significantly higher correlation with human ratings over prior evaluation metrics (\eg, CLIP, BLIP, BLIP2) (Sec.~\ref{sec:vqa-score-results}).

\begin{figure}[t]
\vskip -0.15in
\begin{center}
\centering
     \begin{subfigure}[b]{1.\columnwidth}
         \centering
         \includegraphics[width=1.\textwidth]{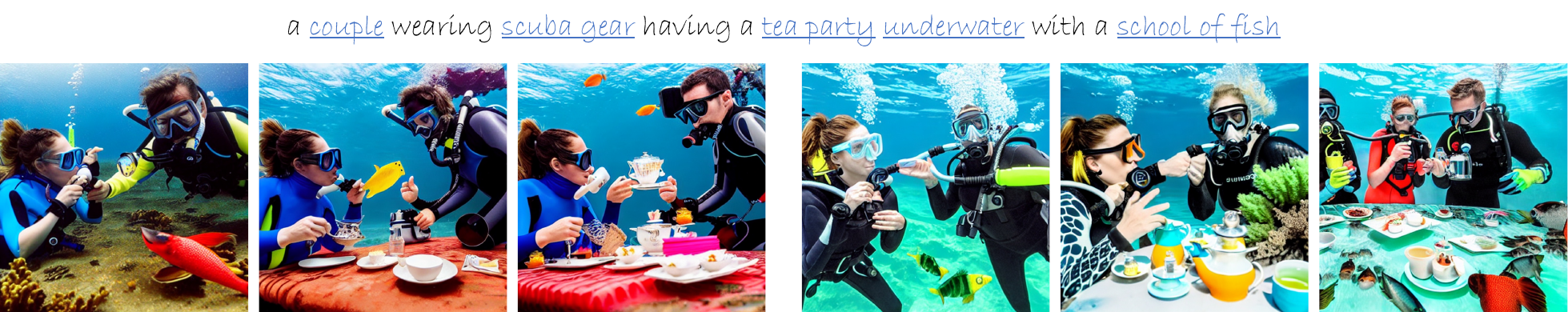}
     \end{subfigure}
     \vskip 0.05in
     \begin{subfigure}[b]{1.\columnwidth}
         \centering
         \includegraphics[width=1.\textwidth]{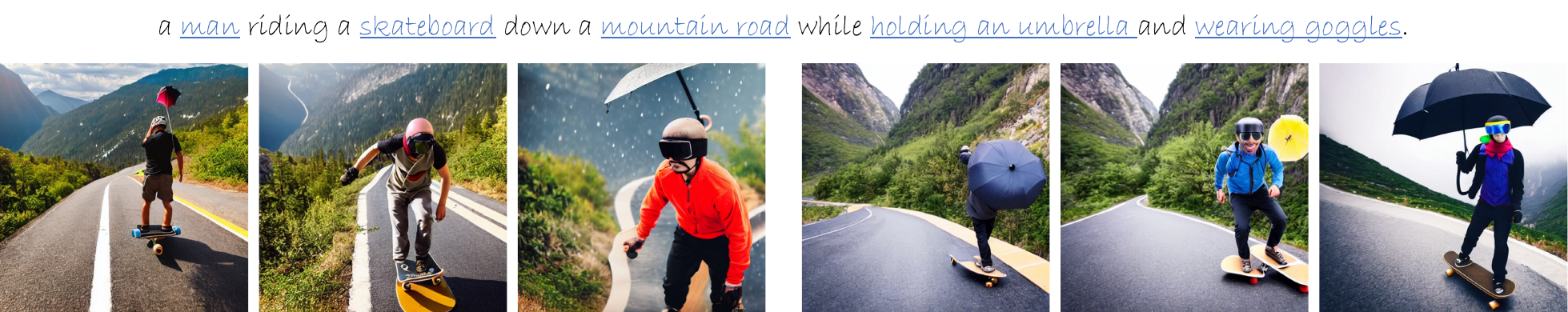}
     \end{subfigure} 
\caption{{\emph{Iterative refinement  (Col:1-3;4-6) for improving text-to-image alignment.}} We propose a simple iterative refinement approach which uses the decompositional alignment scores (refer Sec.~\ref{sec:vqa-score}) as feedback to gradually improve
the alignment of the generated images with the input text-prompt.
}
\label{fig:iter-refinement}
\end{center}
\vskip -0.1in
\end{figure}

Furthermore, we also find that the assertion-level alignment scores provide a useful and explainable feedback for
determining which parts of the input prompt are not being accurately described in the output image. We show that this feedback can then be used to gradually improve the alignment of the generated images with the input text prompt. To this end, we propose a simple iterative refinement procedure (Fig.~\ref{fig:iter-refinement}), wherein at each iteration the expressivity of the least-aligned assertion is improved by increasing the weightage/cross-attention strength (refer Sec.~\ref{sec:vqa-iter}) of corresponding prompt tokens during the reverse diffusion process. 
Through both qualitative and quantitative analysis,
we find that the proposed iterative refinement process allows for generation of better aligned image outputs over prior works \cite{chefer2023attend,liu2022compositional,feng2022training} while 
on average 
showing comparable inference times (Sec.~\ref{sec:vqa-iter-results}).

\section{Related Work}
\label{sec:related-work}

\textbf{Text to Image Generation Models.} Text conditional image synthesis is a topic of keen interest in the vision community. For instance, \cite{xu2018attngan,tao2022df,ye2021improving,zhang2021cross,zhu2019dm} use GANs to perform text guided image generation. Similarly, \cite{ramesh2021zero,yu2022scaling} explore the use of autoregressive models for zero-shot text to image generation. 
Recently, diffusion-based-models \cite{rombach2021highresolution,nichol2021glide,saharia2022photorealistic,ramesh2022hierarchical,yu2022scaling, balaji2022ediffi,zhang2023text} have emerged as a powerful class of methods for performing text-conditional image synthesis over diverse range of target domains.

While remarkable, generating images which align perfectly with the input text-prompt remains a challenging problem \cite{liu2022compositional,feng2022training,chefer2023attend,liu2022character}. To enforce, heavier reliance of generated outputs on the provided text, classifier-free guidance methods \cite{ho2022classifier,nichol2021glide,saharia2022photorealistic} have been proposed. Similarly, use of an additional guidance input to improve controllability of text-to-image generation have recently been extensively explored \cite{meng2022sdedit,choi2021ilvr,singh2022high,avrahami2022spatext,bar2023multidiffusion,sarukkai2023collage,zeng2022scenecomposer,yang2022reco,zhang2023adding,mou2023t2i,ruiz2022dreambooth,gal2022image}. However, even with their application, the generated images are often observed to exhibit fine-grain misalignments such as 
missing secondary objects \cite{chefer2023attend,liu2022compositional} with the input text prompt.

\textbf{Evaluating Image-Text Alignment.} Various protocols for evaluating text-image alignment in a reference-free manner have been proposed \cite{hessel2021clipscore,li2022blip,li2023blip}. Most prior works \cite{hessel2021clipscore, saharia2022photorealistic, nichol2021glide,yu2022scaling} typically use the cosine similarity between the text and image embedding from large-scale multi-modal models \cite{hessel2021clipscore,radford2021learning,yu2022coca,li2021align} such as CLIP \cite{hessel2021clipscore}, BLIP \cite{li2022blip}, BLIP-2 \cite{li2023blip} for evaluating the alignment scores. Recently, \cite{li2022blip,li2023blip} also show the application of BLIP/BLIP-2 models for image-text matching using image retrieval. However, 
as shown in Fig.~\ref{fig:overview}, 
these scores can
give very high scores even if the generated images do not full align with the input text prompt. Furthermore, unlike our approach image-text alignment is often represented through a single scalar value which does not provide an explainable measure which can be used to identify/improve weaknesses of the image generation process.

Recent contemporary work \cite{hu2023tifa} also explores the idea of VQA for evaluating T2I alignment. However, the use of VQA is limited to evaluation alone. In contrast, we further demonstrate how the interpretability of VQA scores can then be used to reliably improve the performance of T2I  models. 

\textbf{Improving Image-Text Alignment.} Recently several works \cite{liu2022compositional,feng2022training,chefer2023attend} have been proposed to explore the problem of improving image-text alignment in a training free manner. Liu \etal \cite{liu2022compositional} propose to modify the reverse diffusion process by composing denoising vectors for different image components.
However, it has been observed \cite{chefer2023attend} that it struggles while generating photorealistic compositions of diverse objects. Feng \etal \cite{feng2022training} use scene graphs to split the input sentence into several noun phrases and then assign a designed attention map to the output of the cross-attention operation. 
In another recent work, Chefer \etal \cite{chefer2023attend} extend the idea of cross-attention map modification to minimize missing objects but instead do so by modifying the noise latents during the reverse diffusion process. While effective at reducing missing objects, we find that the performance / quality of output images can suffer as the number of subjects in the input prompt increases (refer Sec.~\ref{sec:vqa-iter-results}).

Besides training-free methods, recent contemporary work \cite{lee2023aligning,wu2023better} has also explored the possibility of improving image-text alignment using human feedback to finetune existing latent diffusion models. However this often requires the collection of large-scale human evaluation scores and finetuning the diffusion model across a range of diverse data modalities which can be expensive. In contrast, we explore a training free approach for improvement of fine-grain text-to-image alignment.




\section{Our Method}

\begin{figure}[t]
\vskip -0.1in
\begin{center}
    \centerline{\includegraphics[width=1.\linewidth]{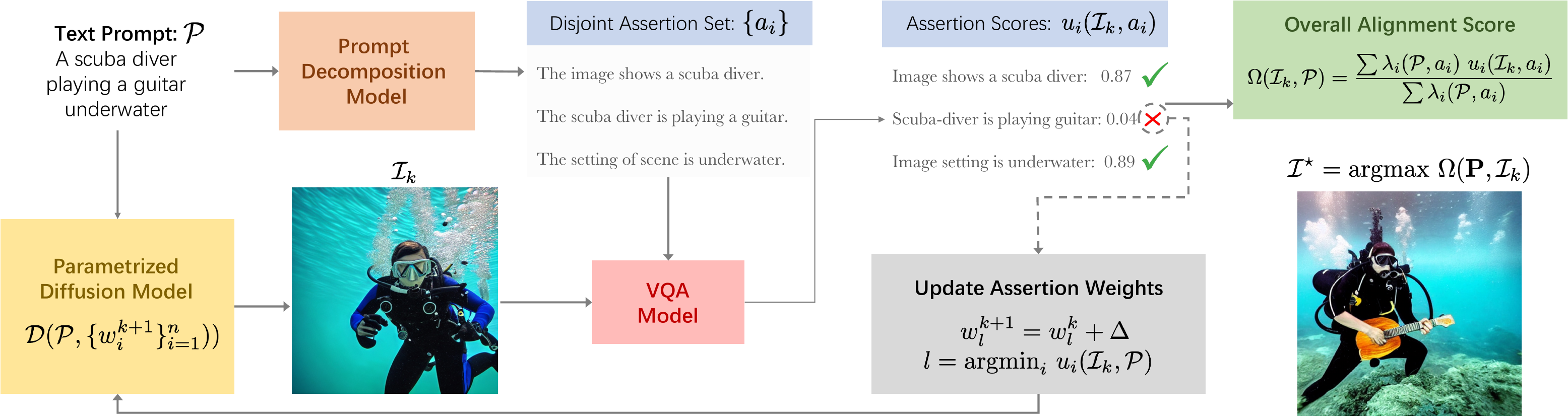}}
    \caption{\emph{{Method Overview.}} Given a text prompt $\mathcal{P}$ and an initially generated output $\mathcal{I}_0$, we first generate a set of disjoint assertions $a_i$ regarding the content of the caption. The alignment of the output image $\mathcal{I}_0$ with each of these assertions is then calculated using a VQA model. Finally, we use the assertion-based-alignment scores $u_i(\mathcal{I}_0,\mathcal{P})$ as feedback to increase the weightage $w_i$ (of the assertion with least alignment score) in a parameterized diffusion model formulation $\mathcal{D}$ (Sec.~\ref{sec:vqa-iter}). This process can then be performed in an iterative manner to gradually improve the quality of the generated outputs until a desirable threshold for the overall alignment score $\Omega(\mathcal{I}_k,\mathcal{P})$ is reached. 
    }
\label{fig:method-overview}
\end{center}
\vskip -0.2in
\end{figure}





Given the image generation output $\mathcal{I}$ corresponding to a text prompt $\mathcal{P}$, we wish to develop a mechanism for evaluation and improvement of fine-grain text-to-image alignment. 
The core idea of our approach is to take a decompositional strategy for both these tasks. 
To this end, we first generate a set of disjoint assertions regarding the content of the input prompt. The alignment of the output image $\mathcal{I}$ with each of these assertions is then calculated using a VQA model. Finally, we use the assertion-based-alignment scores as feedback to improve the 
expressiveness
of the assertion with the least alignment score. This process can then be performed in an iterative manner to gradually improve the quality of  generated outputs 
until a desired value for the overall alignment score is attained.

In the next sections, we discuss each of these steps in detail. In Sec.~\ref{sec:vqa-score} we first discuss the process for evaluating decompositional-alignment scores. We then discuss the iterative refinement process for improving text-to-image alignment in Sec.~\ref{sec:vqa-iter}. Fig.~\ref{fig:method-overview} provides an overview for the overall approach.

\subsection{Evaluating Text-to-Image Alignment}
\label{sec:vqa-score}

\textbf{Prompt Decomposition Model.} 
Given an input prompt $\mathcal{P}$, we first decompose its textual information into a set of disjoint assertions (and corresponding questions) which exhaustively cover the contents of the input prompt. Instead of relying on human-inputs as in \cite{liu2022compositional,chefer2023attend}\footnote{Prior works on improving image-text alignment often rely on human-user inputs for expressing contents of the input prompt into its simpler constituents. For instance, Feng \etal \cite{liu2022compositional} require the user to describe the prompt as a conjunction/disjunction of simpler statements. Similarly, Chefer \etal \cite{chefer2023attend} require the user to provide a set of entities / subjects in the prompt, over which their optimization should be performed.},
we leverage the in-context learning capability \cite{brown2020language} of large-language models \cite{touvron2023llama,liu2023summary} for predicting such decompositions in an autonomous manner. In particular, given an input prompt $\mathcal{P}$ and large-language model $\mathcal{M}$, the prompt decomposition is performed using in-context learning as,
\begin{align}
   \mathbf{x} = \{x_0,x_1, \dots x_n\} = \mathcal{M}(\mathbf{x} \mid \mathcal{P}, D_{exempler},\mathcal{T}),
\end{align}
where $\mathbf{x}$ is the model output, $n$ is the number of decompositions, $D_{exemplar}$ is the in-context learning dataset consisting 4-5 human generated examples for prompt decomposition, and $\mathcal{T}$ is task description. Please refer supp.~material for further details on exemplar-dataset and task-description design.

The model output $\mathbf{x}$ is predicted to contain tuples ${x}_i = \{a_i, p_i\}$, where each tuple is formatted to contain assertions $a_i$ 
and the sub-part $p_i$ of the original prompt $\mathcal{P}$ corresponding to the generated assertion. For instance, given $\mathcal{P}: \textit{`a cat and a dog'}$ the prompt decomposition can be written as,
\begin{align*}
    \mathcal{M}(\mathbf{x} \mid \mathcal{P}: \textit{`a cat and a dog'}, D_{exempler},\mathcal{T}) = \left[ \{ \textit{`there is a cat'},\textit{`a cat'}\} ,   \{\textit{`there is a dog',`a dog'}\}  \right].
\end{align*}

\textbf{Computing Assertion-based Alignment Scores.} 
We next compute the alignment of the generated image $\mathcal{I}$ with each of the disjoint assertions using a Visual-Question-Answering (VQA) model \cite{li2022blip}. In particular, given image $\mathcal{I}$, assertions $a_i, i=1,\dots n$, their rephrasing in question format $a^q_i$ and VQA-model $\mathcal{V}$, the assertion-level alignment scores $u_i(\mathcal{I}, a_i)$ are computed as,
\begin{align*}
    u_i(\mathcal{I}, a_i) = \frac{\exp{(\alpha_i/\tau)}}{\exp{(\alpha_i/\tau)} + \exp{(\beta_i/\tau)}}, \ \textit{where} \quad \alpha_i = \mathcal{V} (\textit{`yes'} \mid \mathcal{I}, a^q_i), \quad  \beta_i = \mathcal{V} (\textit{`no'} \mid \mathcal{I}, a^q_i), 
\end{align*}
where $\alpha_i, \beta_i$ refer to the logit-scores of VQA-model $\mathcal{V}$ for input tuple (image $\mathcal{I}$, question $a^q_i$) corresponding to output tokens \emph{`yes',`no'} respectively. Hyperparameter $\tau$ controls the temperature of the softmax operation and controls the confidence of the alignment predictions.

\begin{figure}[t]
\vskip -0.15in
\begin{center}
\centering
     \begin{subfigure}[b]{1.\columnwidth}
         \centering
         \includegraphics[width=1.\textwidth]{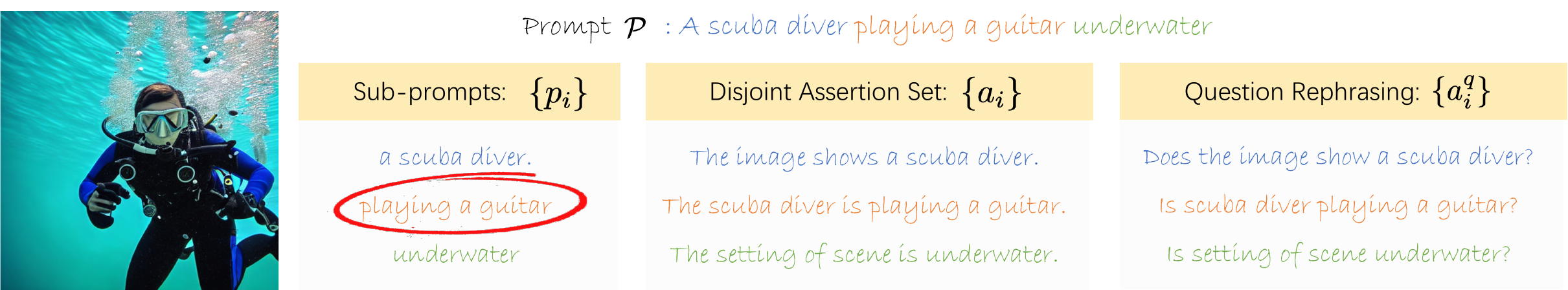}
     \end{subfigure}
     \vskip 0.05in
     \begin{subfigure}[b]{1.\columnwidth}
         \centering
         \includegraphics[width=1.\textwidth]{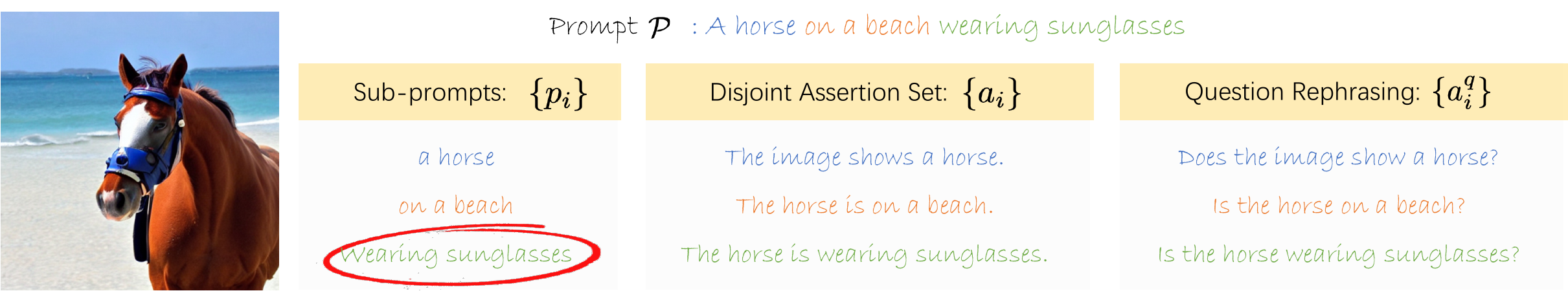}
     \end{subfigure}
     \vskip 0.05in
     \begin{subfigure}[b]{1.\columnwidth}
         \centering
         \includegraphics[width=1.\textwidth]{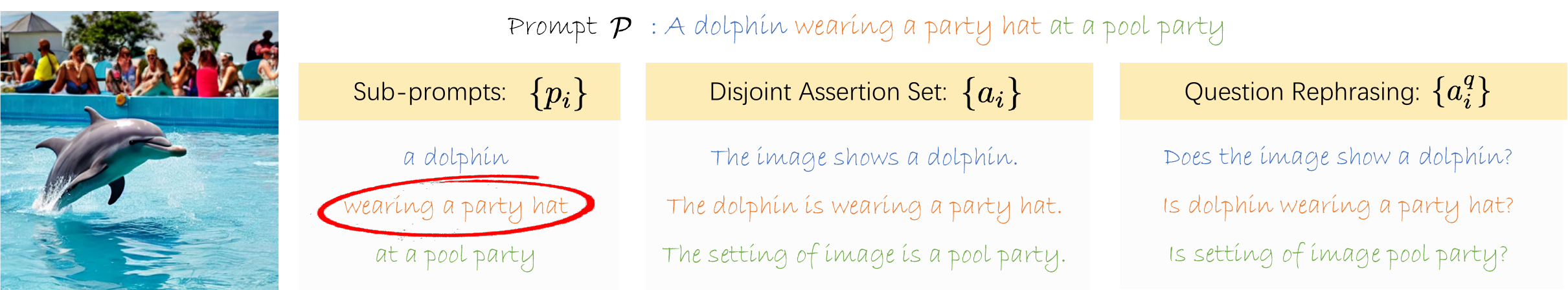}
     \end{subfigure}
\caption{{\emph{ Visualizing the prompt decomposition process.}} 
By dividing a complex prompt $\mathcal{P}$ into a set of disjoint assertions $a_i$, we are able to identify the sub-prompts  $p_i$ (circled) which are not expressed in the image output using VQA, and thereby address them using iterative refinement (Sec.~\ref{sec:vqa-iter}).}
\label{fig:prompt-decomposition}
\end{center}
\vskip -0.1in
\end{figure}



\textbf{Combining Alignment Scores.} Finally, the assertion level alignment-scores $u_i(\mathcal{I}, a_i)$ are combined to give the overall text-to-image alignment score $\Omega(\mathcal{I},\mathcal{P})$ between image $\mathcal{I}$ and prompt $\mathcal{P}$ as,
\begin{align}
    \Omega(\mathcal{I},\mathcal{P}) =  \frac{\sum_i \lambda_i(\mathcal{P},a_i) \ u_i(\mathcal{I}_k, a_i)}{\sum_i \lambda_i(\mathcal{P},a_i)},
\end{align}
where weights $\lambda_i(\mathcal{P},a_i)$ refer to the importance of assertion $a_i$ in capturing the overall content of the input prompt $\mathcal{P}$, and allows the user to control the relative importance of different assertions in generating the final image output\footnote{For simplicity reasons, we mainly use $\lambda_i=1 \forall i$ in the main paper. Further analysis on variable $\lambda_i$ to account for variable information content or visual verifiability of an assertion are provided in supp. material.}. Please refer Fig.~\ref{fig:method-overview} for the overall implementation.

\subsection{Improving Text to Image Alignment}
\label{sec:vqa-iter}

In addition to predicting overall text-to-image alignment score, we find that assertion-level alignment scores $u_i(\mathcal{I}, a_i)$ also provide a useful and explainable way for determining which parts of the input prompt $\mathcal{P}$ are not being accurately described in the output image $\mathcal{I}$. This feedback can then be used in an iterative manner to improve the expressivity of the assertion with least alignment score $u_i(\mathcal{I}, q_i)$, until a desired threshold for the overall text-image alignment score $\Omega(\mathcal{I},\mathcal{P})$ is obtained.




\textbf{Parameterized Diffusion Model}. We first modify the image generation process of standard diffusion models in order to control the expressiveness of different assertions $a_i$ in parametric manner. In particular, we modify the reverse diffusion process to also receive inputs weights $w_i$, where each $w_i$ controls the relative importance of assertion $a_i$ during the image generation process.
In this paper, we mainly consider the following two methods for obtaining such parametric control.


\textbf{Prompt Weighting.} Instead of computing the CLIP \cite{radford2021learning} features from original prompt $\mathcal{P}$ 
we use prompt-weighting \cite{loopback} to modify the input CLIP embeddings to the diffusion model as,
\begin{align}
    \text{CLIP}(\mathcal{P}) = \mathcal{W}(\mathcal{P}, \{\text{CLIP}(p_i),w_i\}_{i=1}^n))
\end{align}
where $\mathcal{W}$ refers to the prompt-weighting function from \cite{rombach2021highresolution,loopback}, $p_i$ refers to the sub-prompt (Sec.~\ref{sec:vqa-score}) corresponding to assertion $a_i$, and weights $w_i$ 
control the relative weight of different sub-prompts $p_i$ in computing the overall CLIP embedding for prompt $\mathcal{P}$.



\textbf{Cross-Attention Control.} 
Similar to \cite{chefer2023attend}, we also explore the idea of modifying the noise latents $z_t$ during the reverse diffusion process, to increase the cross-attention strength of the main noun-subject for each sub-assertion $a_i$. However, instead of only applying the gradient update for the least dominant subject \cite{chefer2023attend}, we modify the loss for the latent update in parametric form as,
\begin{align}
    z_t = z_t - \alpha \nabla_{z_t} \mathcal{L}(z_t, \{w_i\}_{i=1}^n)), \ \mbox{where} \ \quad \mathcal{L}(z_t, \{w_i\}_{i=1}^n) = \sum_i w_i (1- \text{max} \  G(\mathcal{A}^t_i)),
\end{align}
where $\alpha$ is the step-size, $\mathcal{A}^t_i$ refer to the attention map corresponding to the main noun-subject in assertion $a_i$, $G$ is a smoothing function and weights $w_i$ control the extent to which the expression of different noun-subjects in the prompt (for each assertion) will be increased in the next iteration.

\textbf{Iterative Refinement.} Given the above parametric formulation for controlling expression of different assertions, we next propose a simple yet effective iterative refinement approach towards improving text-to-image alignment. In particular, at any iteration $k \in [1,5]$ during the refinement process, we first compute both overall text-image similarity score $\Omega(\mathcal{I}_k,\mathcal{P})$ and assertion-level alignment scores $u_i(\mathcal{I}_k,\mathcal{P})$.
The image generation output $\mathcal{I}_{k+1}$ for the next iteration is then computed as,
\begin{align}
    \mathcal{I}_{k+1} = \mathcal{D}(\mathcal{P},\{w^{k+1}_i\}_{i=1}^n)); \ \mbox{where} \quad  
    w_i^{k+1} = 
    \begin{cases}
    w_i^k + \Delta, \ \text{if} \quad  i = \text{argmin}_{l}  \ u_l(\mathcal{I},\mathcal{P})\\
    w_i^k \quad \text{otherwise}
    \end{cases},
\end{align}
where $\mathcal{D}$ refers to the parametrized diffusion model and $\Delta$ is a hyper-parameter. This iterative process is then performed until a desirable threshold for the overall alignment score $\Omega(\mathcal{I}_k,\mathcal{P})$ is reached. The image generation output $\mathcal{I}^\star$ at the end of the refinement process is then computed as,
\begin{align}
    \mathcal{I}^\star = \text{argmax}_{\mathcal{I}_k} \Omega(\mathcal{I}_k,\mathcal{P}).
\end{align}
\section{Experiments}
\label{sec:experiments}

\begin{figure}[t]
\vskip -0.1in
\begin{center}
    \centerline{\includegraphics[width=1.\linewidth]{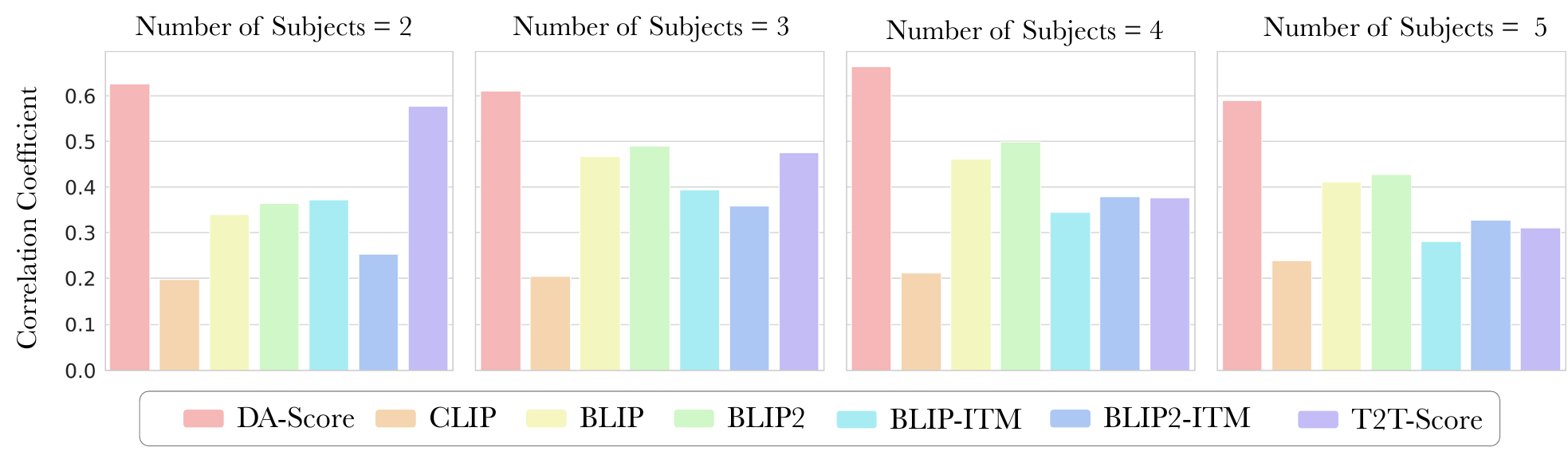}}
    \vskip -0.05in
    \caption{\emph{{Method comparisons w.r.t correlation with human ratings.}} We compare the correlation of different text-to-image alignment scores with those obtained from human subjects, as the number of subjects in the input prompt (refer Sec.~\ref{sec:experiments}) is varied. We observe that the proposed alignment score (DA-score) provides a better match for human-ratings over traditional text-to-image alignment scores.
    }
\label{fig:corr-human}
\end{center}
\vskip -0.2in
\end{figure}

\begin{figure}[t]
\vskip -0.1in
\begin{center}
    \centerline{\includegraphics[width=1.\linewidth]{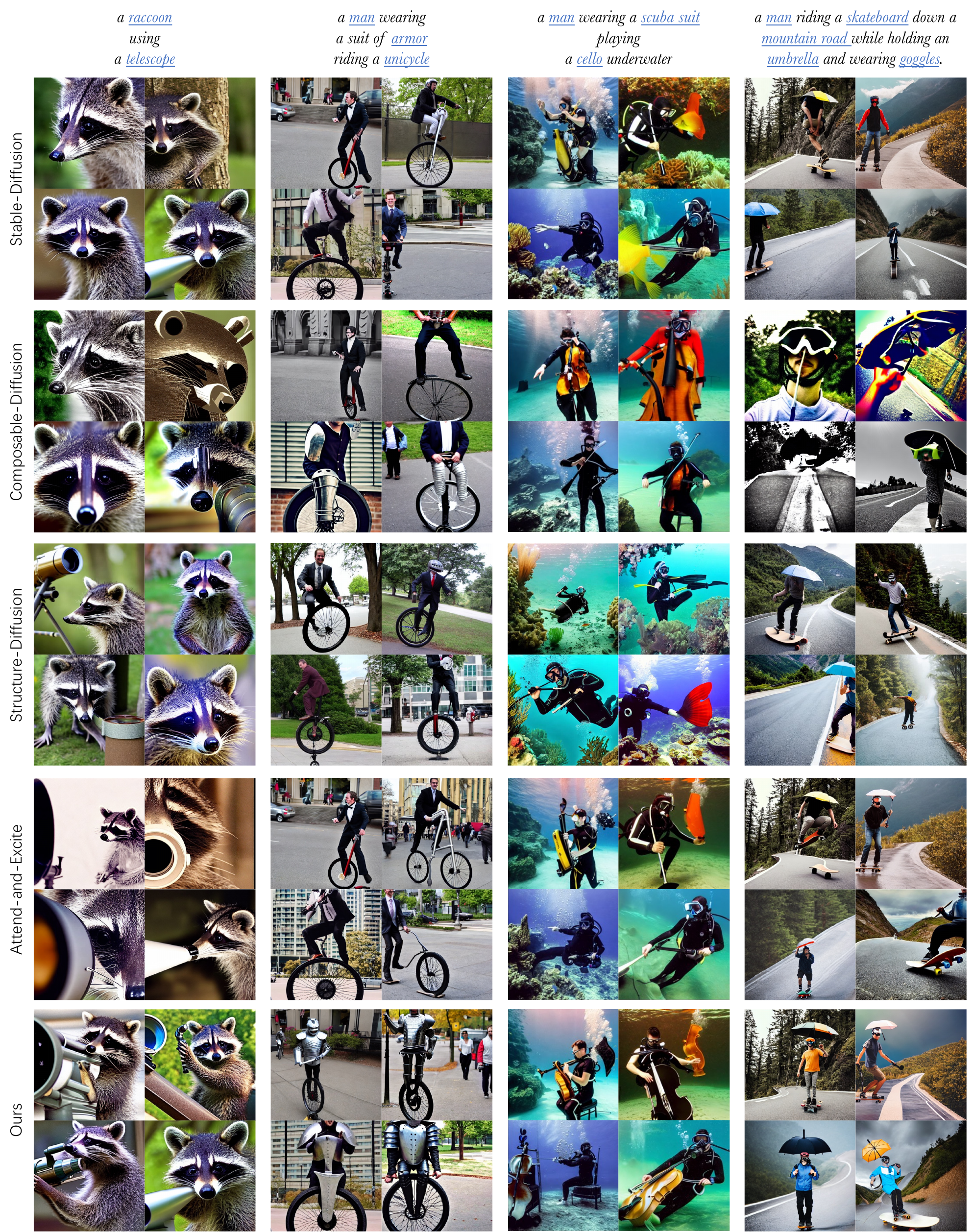}}
\vskip -0.05in
\caption{\emph{{Qualitative comparison w.r.t text-to-image alignment.}} We compare the outputs of our iterative refinement approach with prior works \cite{rombach2021highresolution,liu2022compositional,feng2022training,chefer2023attend} on improving quality of generated images with changing number of subjects (underlined) from 2 to 5. Please zoom-in for best comparisons. }
\label{fig:image-alignment-qual}
\end{center}
\vskip -0.3in
\end{figure}

\begin{figure}[h!]
\begin{center}
\centering
     \begin{subfigure}[b]{1.\columnwidth}
         \centering
         \includegraphics[width=1.\textwidth]{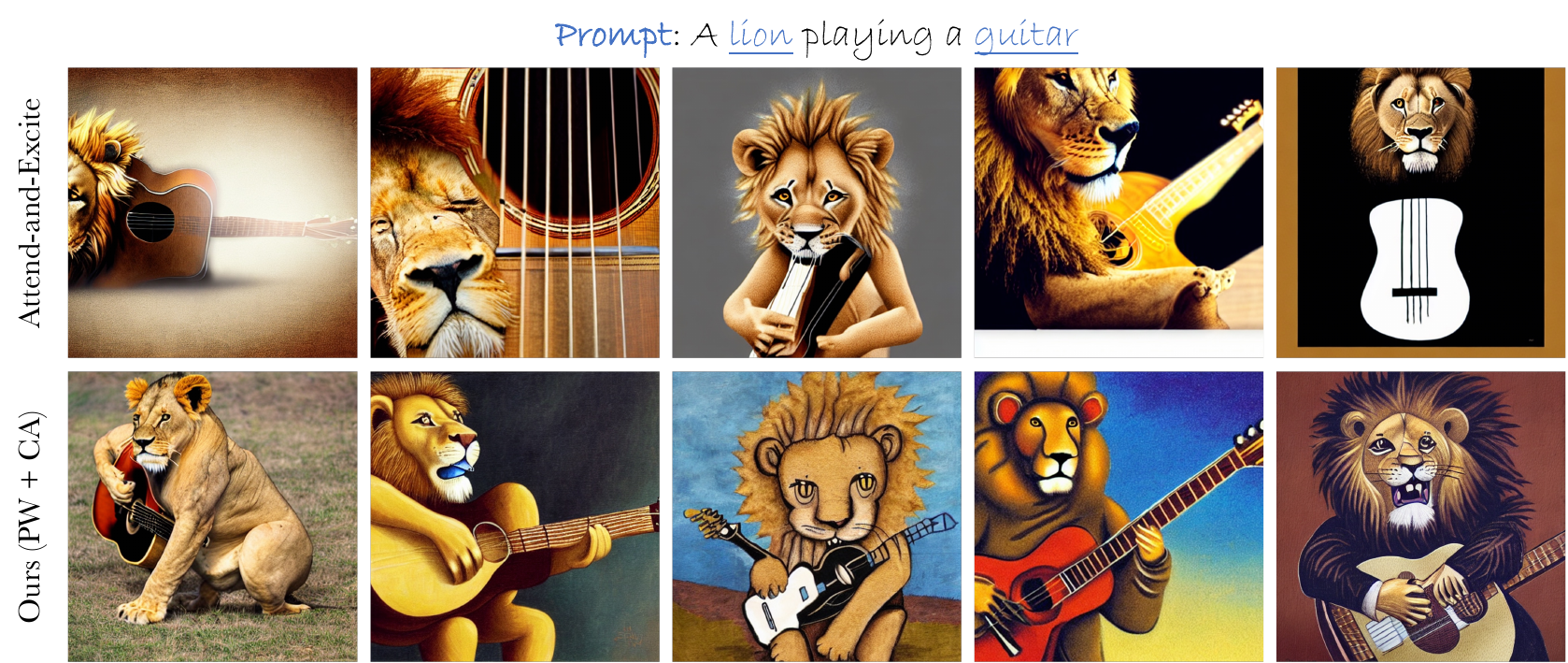}
         \caption{\emph{Object Relationship}: Eval-and-Refine helps better capture both presence and relationship between the objects.}
     \end{subfigure}
     \vskip 0.02in
     \begin{subfigure}[b]{1.\columnwidth}
         \centering
         \includegraphics[width=1.\textwidth]{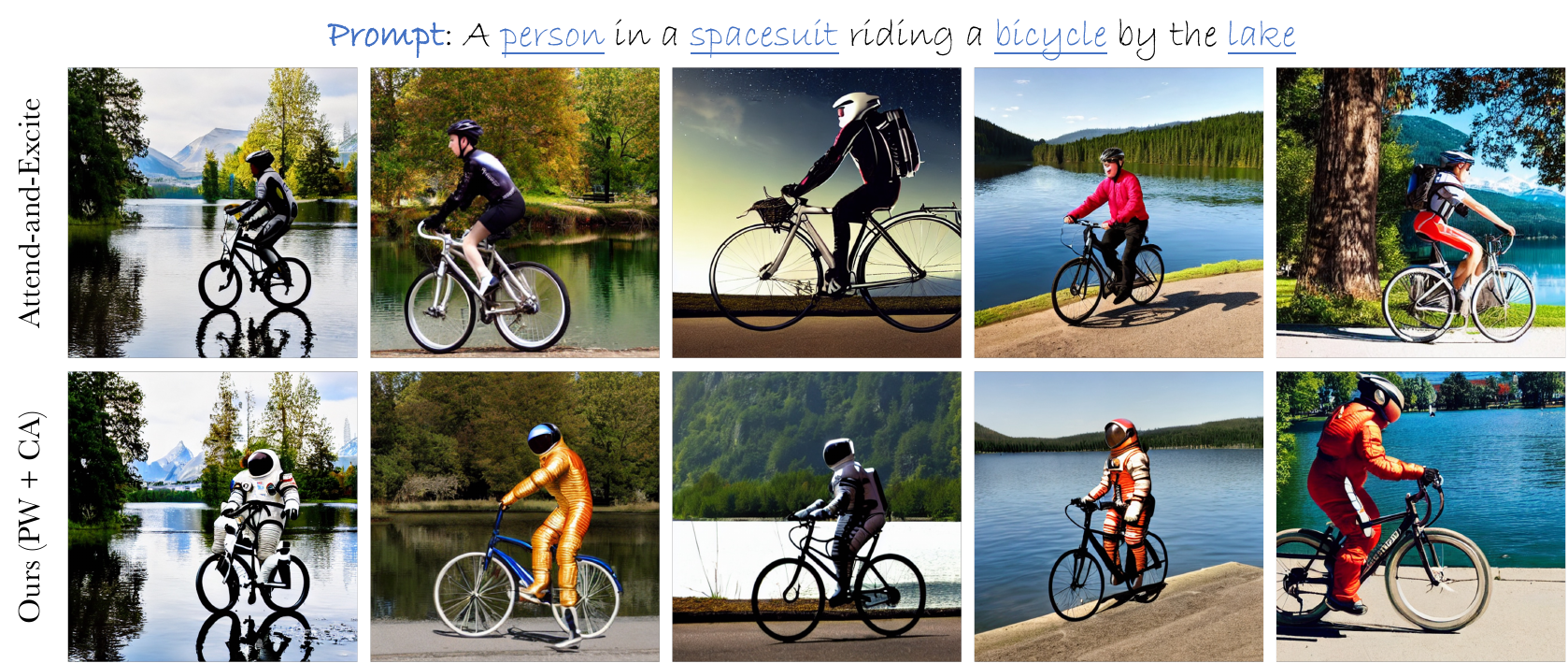}
         \caption{\emph{Overlapping entities}: Proposed approach can better handle cases with overlapping entities (spacesuit, person).}
     \end{subfigure} 
     \vskip 0.02in
     \begin{subfigure}[b]{1.\columnwidth}
         \centering
         \includegraphics[width=1.\textwidth]{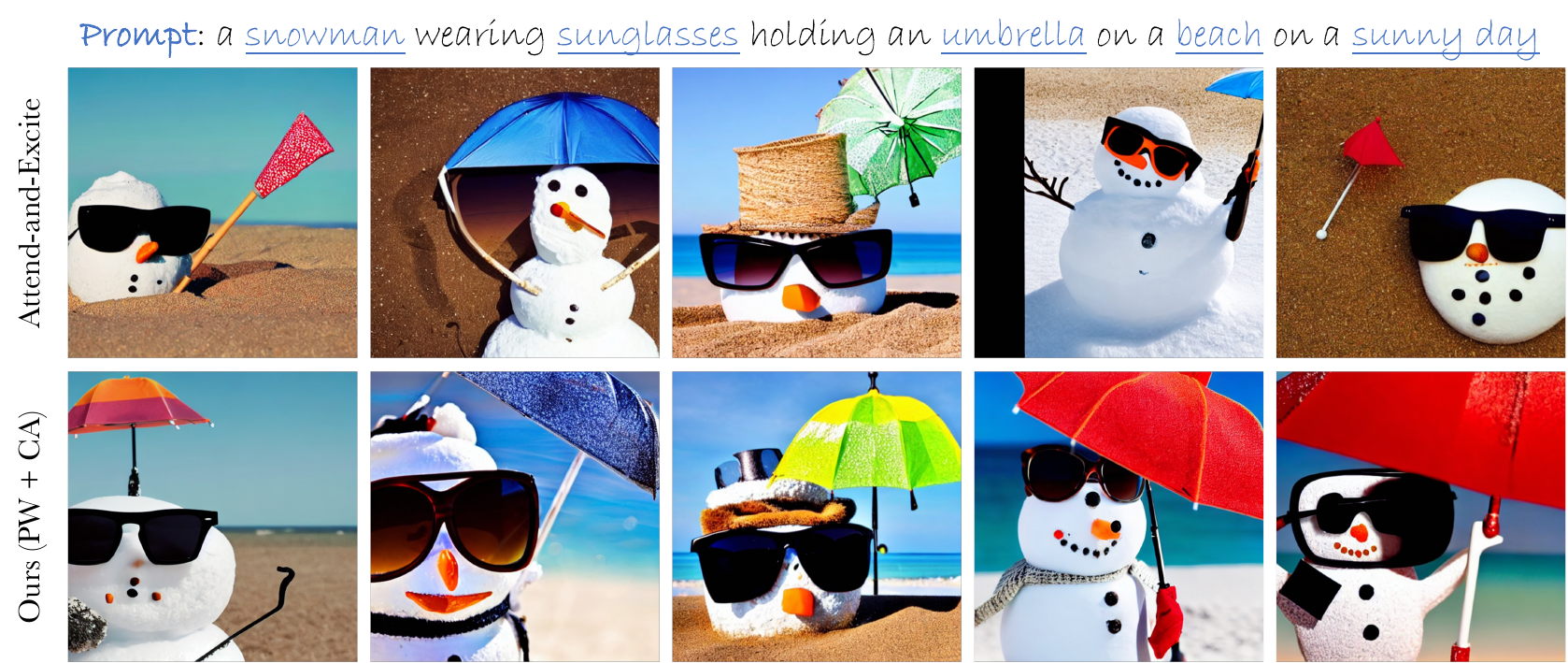}
         \caption{\emph{Prompt Complexity}: Eval-and-Refine shows better alignment as number of subjects in input prompt increase.}
     \end{subfigure} 
\caption{\emph{{Additional comparisons with Attend-and-Excite.}} We analyse three main ways in which the proposed iterative-refinement improves over Attend-and-Excite \cite{chefer2023attend} (refer Sec.~\ref{sec:vqa-iter-results} for details).}
\label{fig:attend-excite-comp}
\end{center}
\vskip -0.1in
\end{figure}

\textbf{Dataset.}
Since there are no openly available datasets addressing semantic challenges in text-based image generation with human annotations, we introduce a new benchmark dataset Decomposable-Captions-4k for method comparison. The dataset consists an overall of 24960 human annotations on images generated using all methods \cite{rombach2021highresolution,liu2022compositional,chefer2023attend} (including ours) across a diverse set of 4160 input prompts.  Each image is a given rating between 1 and 5 (where 1 represents that \textit{`image is irrelevant to the prompt'} and 5 represents that \textit{`image is an accurate match for the prompt'}). 

Furthermore, unlike prior works \cite{chefer2023attend} which predominantly analyse the performance on relatively simple prompts with two subjects (\eg object a and object b), we construct a systematically diverse pool of input prompts for better understanding text-to-image alignment across varying complexities in the text prompt. In particular, the prompts for the dataset are designed to encapsulate two axis of complexity: \emph{number of subjects} and \emph{realism}. The number of subjects refers to the number of main objects described in the input prompt and varies from 2 (\eg, \emph{a \bline{cat} with a \bline{ball}}) to 5 (\eg, \emph{a \bline{woman} walking her \bline{dog} on a \bline{leash} by the \bline{beach} during \bline{sunset}}). Similarly, the \emph{realism} of a prompt is defined as the degree to which different concepts naturally co-occur together and varies as \emph{easy}, \emph{medium}, \emph{hard} and  \emph{very hard}. \emph{easy} typically refers to prompts where concepts are naturally co-occurring together (\eg, \emph{a \bline{dog} in a \bline{park}}) while \emph{very hard} refers to prompts where concept combination is very rare (\eg, \emph{a \bline{dog} playing a \bline{piano}}). Further details regarding the dataset are provided in supplementary material.


\subsection{Evaluating Text-to-Image Alignment}
\label{sec:vqa-score-results}

\textbf{Baselines.} We compare the performance of the \emph{Decompositional-Alignment Score} with prior works on evaluating text-to-image alignment in a reference-free manner. In particular, we show comparisons with CLIP \cite{hessel2021clipscore}, BLIP \cite{li2022blip} and BLIP2 \cite{li2023blip} scores where the text-to-image alignment score is computed using the cosine similarity between the corresponding image and text embeddings. We also include comparisons with BLIP-ITM and BLIP2-ITM which directly predict a binary image-text matching score (between 0 and 1) for input prompt and output image. 
Finally, we report results on the recently proposed text-to-text (T2T) similarity metric \cite{chefer2023attend} which computes image-text similarity as the average cosine similarity between input prompt and captions generated (using BLIP) from the input image. 

\textbf{Quantitative Results.} Fig.~\ref{fig:corr-human} shows the correlation between human annotations and predicted text-to-image alignment scores across different metrics on the \emph{Decomposable-Captions} dataset. We observe that the \emph{DA-Score} shows a significantly higher correlation with human evaluation ratings as opposed to prior works across varying number of subjects $N \in [2,5]$ in the input prompt. We also note that while the recently proposed T2T similarity score \cite{chefer2023attend} shows comparable correlation with ours for $N=2$,
its performance significantly drops as the number of subjects in the input prompt increases. 


\subsection{Improving Text-to-Image Alignment}
\label{sec:vqa-iter-results}

\begin{figure}[t]
\vskip -0.1in
\begin{center}
    \centerline{\includegraphics[width=1.\linewidth]{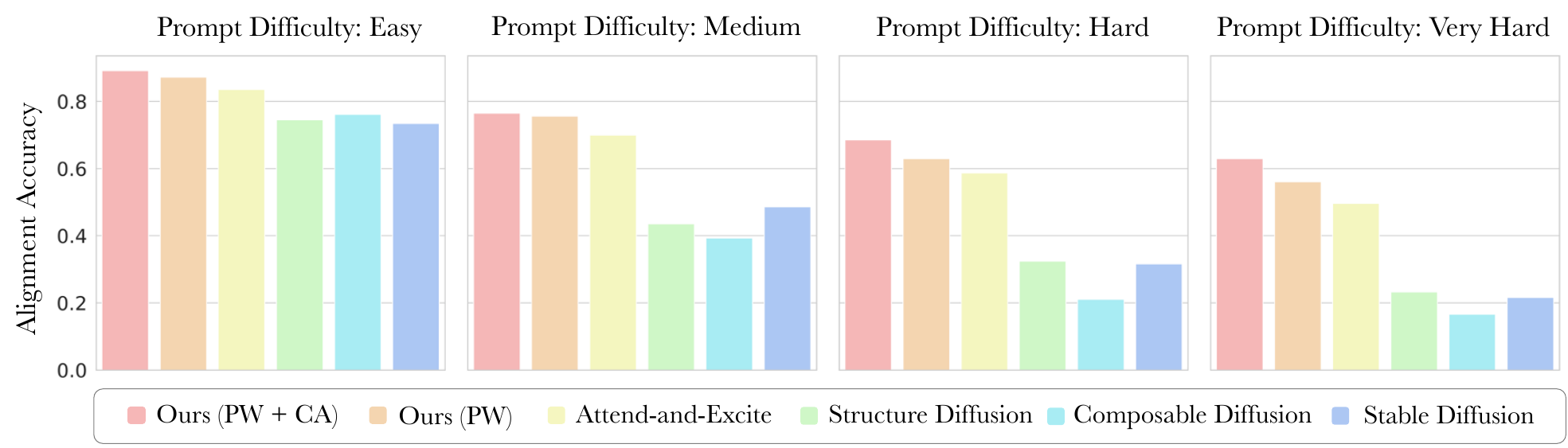}}
    \vskip -0.05in
     \caption{\emph{{Variation of alignment accuracy with prompt difficulty.}} We observe that while the accuracy of all methods decreases with increasing difficulty in prompt \emph{realism } (refer Sec.~\ref{sec:experiments}), the proposed iterative refinement approach consistently performs better than prior works.}
\label{fig:acc-var}
\end{center}
\vskip -0.1in
\end{figure}

In this section, we compare the performance of our iterative refinement approach with prior works on improving text-to-image alignment in a training-free manner. In particular, we show comparisons with 1) Stable Diffusion \cite{rombach2021highresolution}, 2) Composable Diffusion \cite{liu2022compositional} 3) StructureDiffusion \cite{feng2022training} and 4) Attend-and-Excite \cite{chefer2023attend}.
All images are generated using the same seed across all methods.

\textbf{Qualitative Results.} Results are shown in Fig.~\ref{fig:image-alignment-qual}. As shown, we observe that Composable Diffusion \cite{liu2022compositional} struggles to generate photorealistic combinations of objects especially as number of subjects in the prompt increase. 
StructureDiffusion \cite{feng2022training} helps in addressing some missing objects \eg, telescope in example-1, but the generated images tend to be semantically similar to those produced by the original Stable Diffusion model, and thus does not significantly improve text-to-image alignment. 

Attend-and-Excite \cite{chefer2023attend} shows much better performance in addressing missing objects (\eg, telescope in example-1 and umbrella in example-4). However, as sumamrized in Fig.~\ref{fig:attend-excite-comp} we observe that it suffers from 3 main challenges: 1) \emph{Object Relationship} (Fig.~\ref{fig:attend-excite-comp}a): we observe that despite having desired objects, generated images may sometimes fail to convey relationship between them. For \eg, in row-1 Fig.~\ref{fig:attend-excite-comp} while output images show both a \emph{\bline{lion} and \bline{guitar}}, the lion does not seem to be playing the guitar. In contrast, Eval-and-Refine is able to describe both presence and relation between objects in a better manner. 2) \emph{Overlapping Entities} (Fig.~\ref{fig:attend-excite-comp}b): 
For images with overlapping  entities (\eg, \emph{\bline{person} and \bline{spacesuit}}), we observe that Attend-and-Excite \cite{chefer2023attend} typically spends most of gradient updates balancing between the overlapping entities, as both entities (\emph{\bline{person} and \bline{spacesuit}}) occupy the same cross-attention region. This can lead to outputs  where a) other important aspects (\eg, \emph{\bline{lake}} in Col-3) or b) one of the two entities (\eg, \emph{\bline{spacesuit}}) are ignored.
3) \emph{Prompt Complexity} (Fig.~\ref{fig:attend-excite-comp}c): Finally, we note that since Attend-and-Excite \cite{chefer2023attend} is limited to applying the cross-attention update \emph{w.r.t} the least dominant subject, as the complexity of input prompt $\mathcal{P}$ increases, it may miss some objects (\eg, \emph{\bline{umbrella, beach, sunny day}}) during the generation process. In contrast, the iterative nature of our approach allows it to keep refining the output image $\mathcal{I}$ until a desirable threshold for the overall image-text alignment score $\Omega(\mathcal{I},\mathcal{P})$ is reached.

\textbf{Quantitative Results.} In addition to qualitative experiments, we also evaluate the efficacy of our approach using human evaluations. In this regard, we report three metrics: 1) \emph{normalized human score}: which refers to the average human rating (normalized between 0-1) for images generated on the Decomposable-Captions-4k dataset.
2) \emph{accuracy}: indicating the percentage of generated images which are considered as an accurate match (rating: 5) for the input text prompt by a human subject. 
3) \emph{pairwise-preference}: where human subjects are shown pair of images generated using our method and prior work, and are supposed to classify each image-pair as a win, loss or tie (win meaning our method is preferred).
For our approach we consider two variants 1) \emph{Ours (PW)} which performs iterative refinement using only prompt-weighting, and 2) \emph{Ours (PW + CA)} where iterative refinement is performed using both prompt weighting and introducing cross-attention updates (Sec.~\ref{sec:vqa-iter}). Pairwise preference scores are reported while using \emph{Ours (PW + CA)} while comparing with prior works.

Results are shown in Fig.~\ref{fig:acc-var} and Tab.~\ref{tab:quant-results}.
We observe that while the text-to-image alignment accuracy for all methods decreases with an increased difficulty in input text prompts (Fig.~\ref{fig:acc-var}), we find that the our approach with only prompt-weighting is able to consistently perform on-par or better than Attend-and-Excite \cite{chefer2023attend}.  Further introduction of cross-attention updates (Sec.~\ref{sec:vqa-iter}), allows our approach to exhibit even better performance, which outperforms Attend-and-Excite \cite{chefer2023attend} by  8.67 \% in terms of overall alignment accuracy of the generated images. These improvements are also reflected in the pairwise comparisons where human subjects tend to prefer our approach over prior works \cite{chefer2023attend,liu2022compositional,feng2022training}.

\begingroup
\begin{table}[t]
\begin{center}
\small
\begin{tabular}{l|c|c|ccc|c}
\toprule
\multirow{2}{*}{Method} & Norm.~Human & Alignment & \multicolumn{3}{c|}{Pairwise Comparison \%} & Inference \\
&  Score (\%)  & Accuracy (\%) & Win $\uparrow$ & Tie & Lose $\downarrow$ & Time (s) \\
\midrule
Stable-Diffusion \cite{rombach2021highresolution} & 72.98 & 43.66  & 41.7 & 50.1 & 8.2 & 3.54 s \\ 
Composable-Diffusion \cite{liu2022compositional} & 70.28 & 37.72  & 57.1 & 38.5 & 4.4 & 10.89 s \\ 
Structure-Diffusion \cite{feng2022training} & 74.93 & 45.23  & 37.5 & 54.6 & 7.9  & 11.51 s \\ 
Attend-and-Excite \cite{chefer2023attend} & 85.94 & 65.50  & 23.6 & 62.3 & 14.1 & 8.59 s \\ 
\rowcolor{lightyellow} Ours (PW)  & 89.53 & 70.28  & \cellcolor{Gray}N/A & \cellcolor{Gray}N/A & \cellcolor{Gray}N/A & 10.32 s \\ 
\rowcolor{lightyellow} Ours (PW + CA) & \textbf{90.25} & \textbf{74.16}   & \cellcolor{Gray}N/A & \cellcolor{Gray}N/A & \cellcolor{Gray}N/A & 12.24 s \\ 
\bottomrule
\end{tabular}
\end{center}
\caption{\emph{{Quantitative Results}}. We report text-to-image alignment comparisons \emph{w.r.t}  normalized human rating score ({Col:2}), average alignment accuracy evaluated by human subjects ({Col:3}) and pairwise user-preference scores (ours vs prior work) (Col:4-6). Finally, we also report average inference time per image for different methods in Col:7. We observe that our approach shows better text-to-image alignment performance while on average using marginally higher inference time.}
\label{tab:quant-results}
\vskip -0.15in
\end{table}
\endgroup

\textbf{Inference time comparison.}
Tab.~\ref{tab:quant-results} shows comparison for the average inference time (per image) for our approach with prior works \cite{chefer2023attend,feng2022training,liu2022compositional}. We observe that despite the use of an iterative process for our approach, the overall inference time is comparable with prior works. This occurs because prior works themselves often include additional steps. For instance, Composable-Diffusion \cite{liu2022compositional} requires the computation of separate denoising latents for each statement in the confunction/disjunction operation, thereby increasing the overall inference time almost linearly with number of subjects. Similarly, Attend-and-Excite \cite{chefer2023attend} includes additional gradient descent steps for modifying cross-attention maps. 
Moreover, such an increase is accumulated even if the baseline Stable-Diffusion \cite{rombach2021highresolution} model already generates accurate images. In contrast, the proposed iterative refinement approach is able to adaptively adjust the number of iterations required for the generation process by monitoring the proposed \emph{DA- Score} for evaluating whether the generation outputs are already good enough.

\section{Conclusion}
In this paper, we explore a simple yet effective decompositional approach for both evaluation and improvement of text-to-image alignment with latent diffusion models. To this end, we first propose a \emph{Decompositional-Alignment Score} which given a complex prompt breaks it down into a set of disjoint assertions. The alignment of each of these assertions with the generated image is then measured using a VQA model. The assertion-based alignment scores are finally combined to a give an overall text-to-image alignment score. Experimental results show that proposed metric shows significantly higher correlation with human subject ratings over traditional CLIP, BLIP based image-text matching scores. Finally, we propose a simple iterative refinement approach which uses the decompositional-alignment scores as feedback to gradually improve the quality of the generated images. Despite its simplicity, we find that the proposed approach is able to surpass previous state-of-the-art on text-to-image alignment accuracy while on average using only marginally higher inference times. 
We hope that our research can open new avenues for robust deployment of text-to-image models for practical applications.



\newpage
{
\small
\bibliographystyle{unsrt}
\bibliography{evaluate-and-refine}
}

\newpage
\appendix

\title{Divide, Evaluate, and Refine:  Evaluating and Improving Text-to-Image Alignment with \\Iterative VQA Feedback}
\author{}

\maketitles

\vspace{-2pt}
\section{Additional Results}
\label{sec:additional-results}

In this section, we include additional results for our approach which could not be included in the main paper due to space constraints. In particular, we report additional results for visualizing the iterative refinement process in Sec.~\ref{sec:iter-refinement}. We also provide additional results comparing our method performance with Attend-and-Excite \cite{chefer2023attend} in Sec.~\ref{sec:comp-attend-excite}. Finally, in Sec.~\ref{sec:comp-hps}, we compare our approach with recent contemporary work on using human feedback for improving text-to-image alignment.

\subsection{Visualizing the Iterative Refinement Process}

\label{sec:iter-refinement}

\begin{figure}[h!]
\vskip -0.15in
\begin{center}
\centering
     \begin{subfigure}[b]{1.\columnwidth}
         \centering
         \includegraphics[width=1.\textwidth]{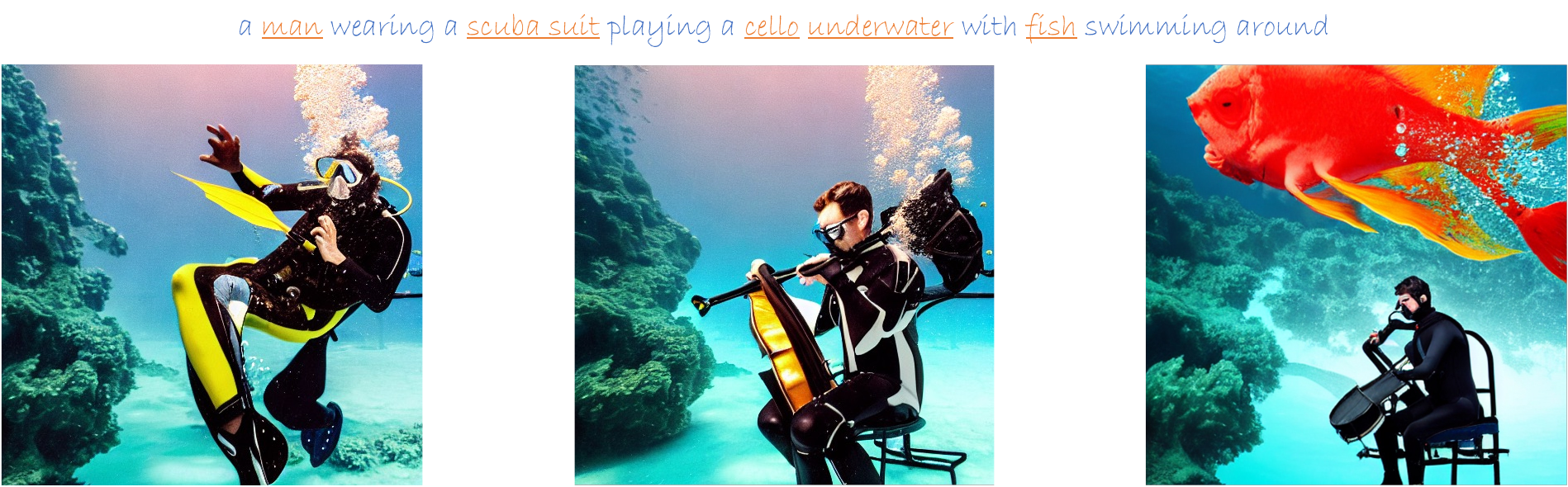}
     \end{subfigure}
     \begin{subfigure}[b]{1.\columnwidth}
         \centering
         \includegraphics[width=1.\textwidth]{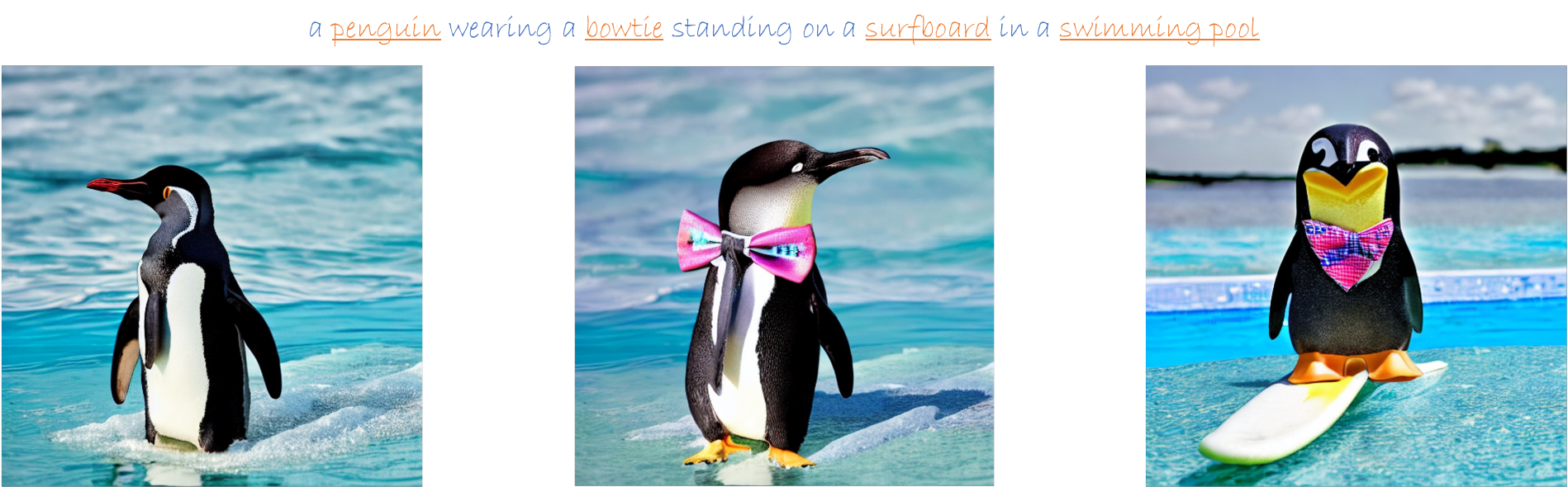}
     \end{subfigure}
     \begin{subfigure}[b]{1.\columnwidth}
         \centering
         \includegraphics[width=1.\textwidth]{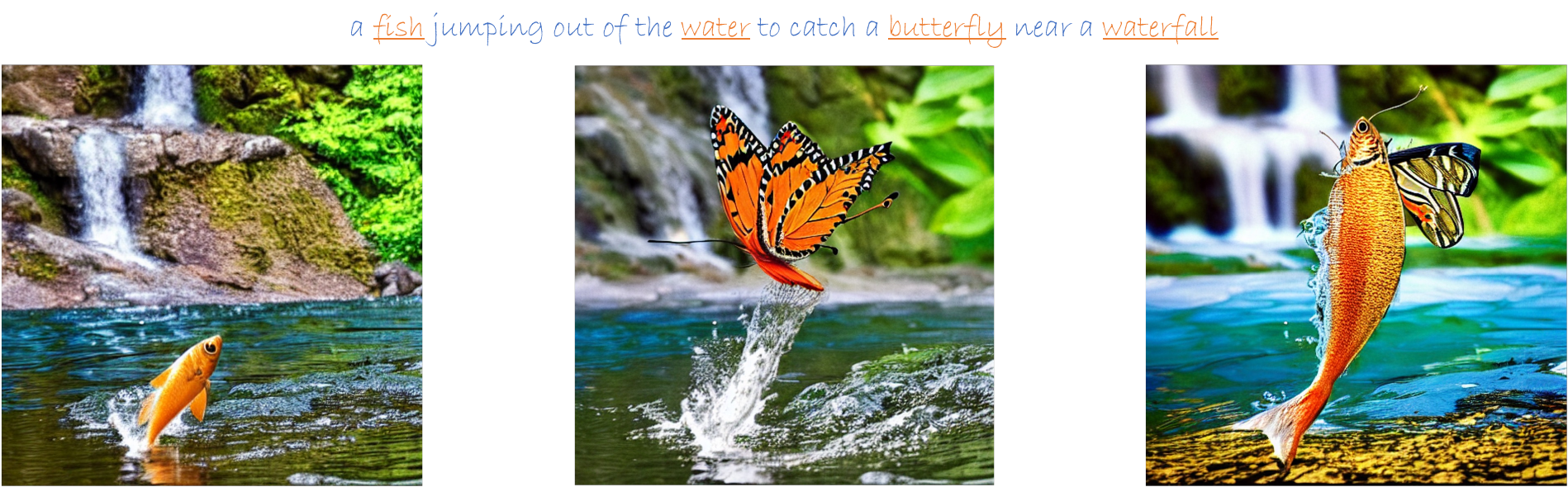}
     \end{subfigure}
\caption{{\emph{ Visualizing iterative refinement process for improving text-to-image alignment.}} 
}
\label{fig:iter-refinement-v1}
\end{center}
\vskip -0.1in
\end{figure}

\begin{figure}[h!]
\vskip -0.1in
\begin{center}
\centering
     \begin{subfigure}[b]{1.\columnwidth}
         \centering
         \includegraphics[width=1.\textwidth]{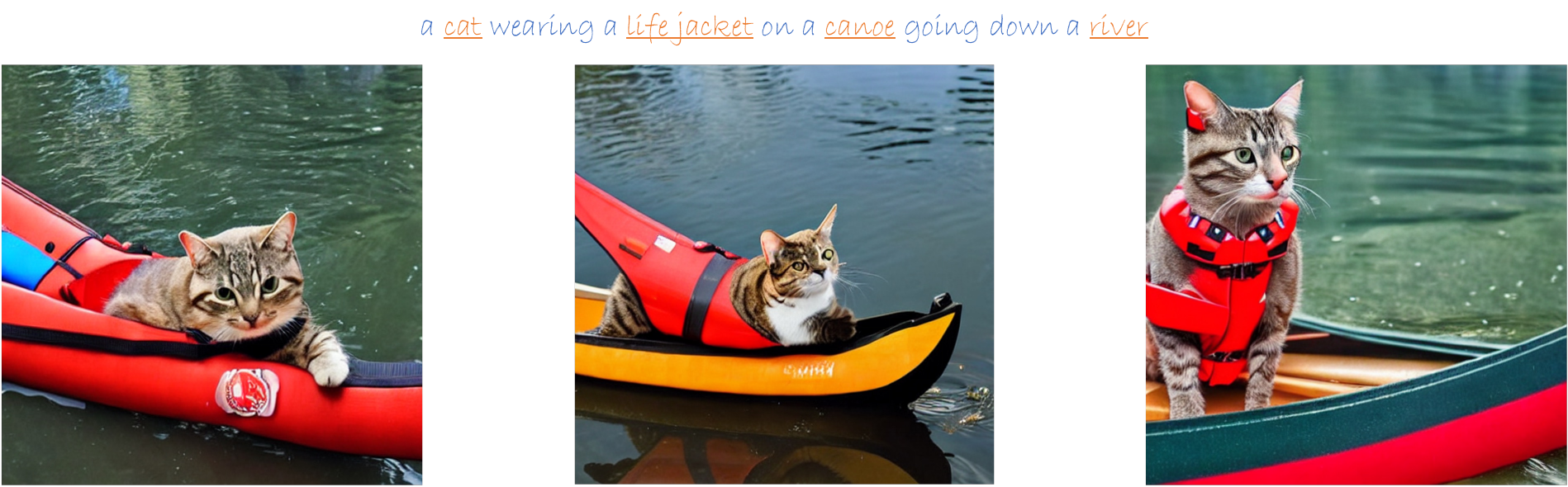}
     \end{subfigure}
     \vskip 0.02in
     \begin{subfigure}[b]{1.\columnwidth}
         \centering
         \includegraphics[width=1.\textwidth]{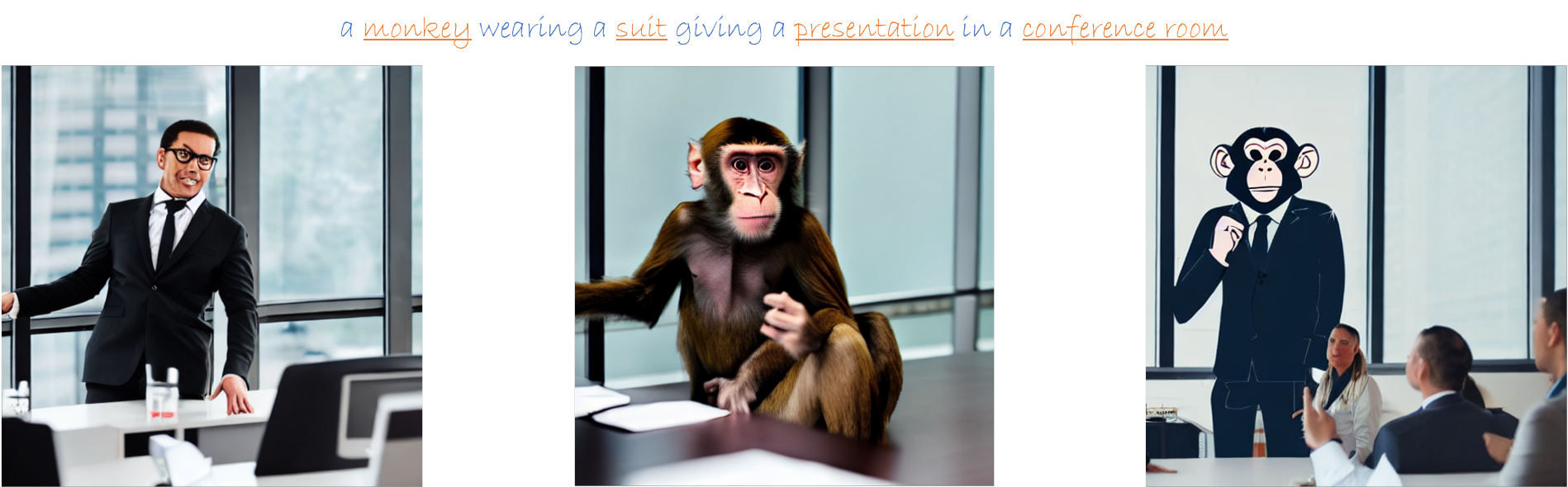}
     \end{subfigure}
     \vskip 0.02in
     \begin{subfigure}[b]{1.\columnwidth}
         \centering
         \includegraphics[width=1.\textwidth]{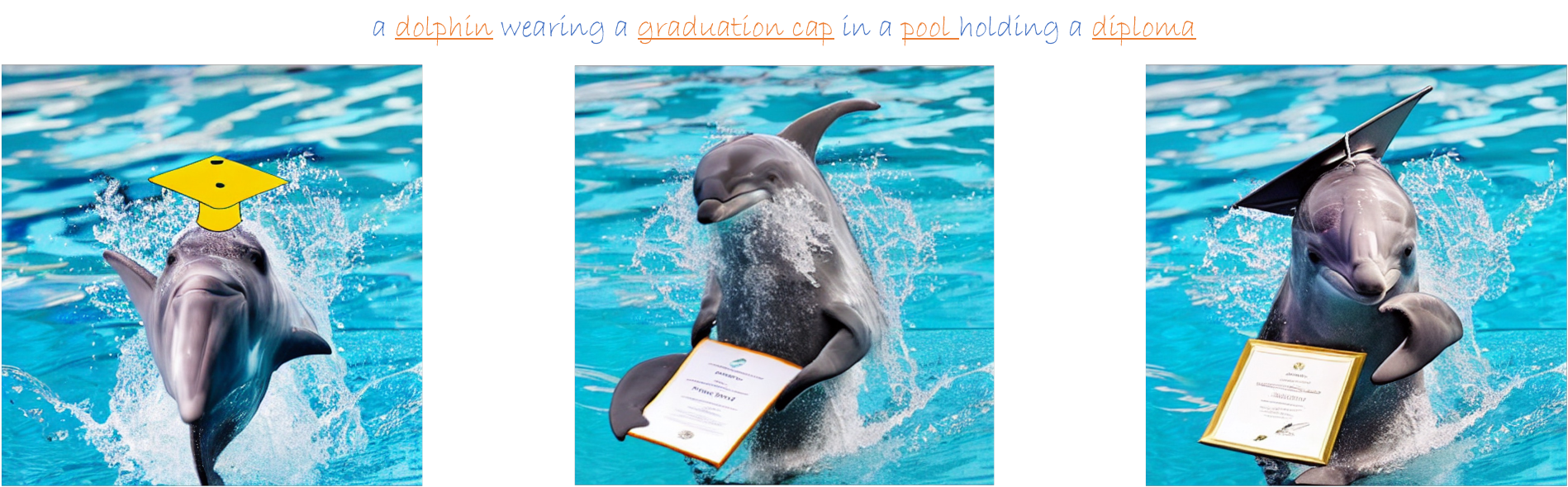}
     \end{subfigure}
     \vskip 0.02in
     \begin{subfigure}[b]{1.\columnwidth}
         \centering
         \includegraphics[width=1.\textwidth]{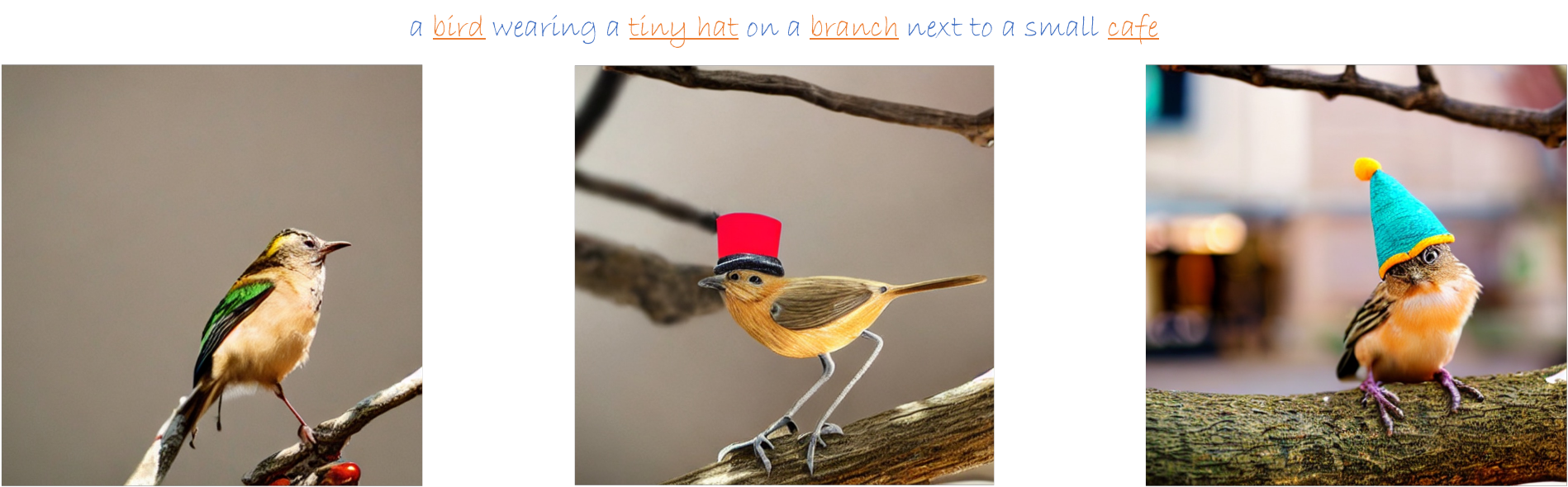}
     \end{subfigure}
      \vskip 0.02in
     \begin{subfigure}[b]{1.\columnwidth}
         \centering
         \includegraphics[width=1.\textwidth]{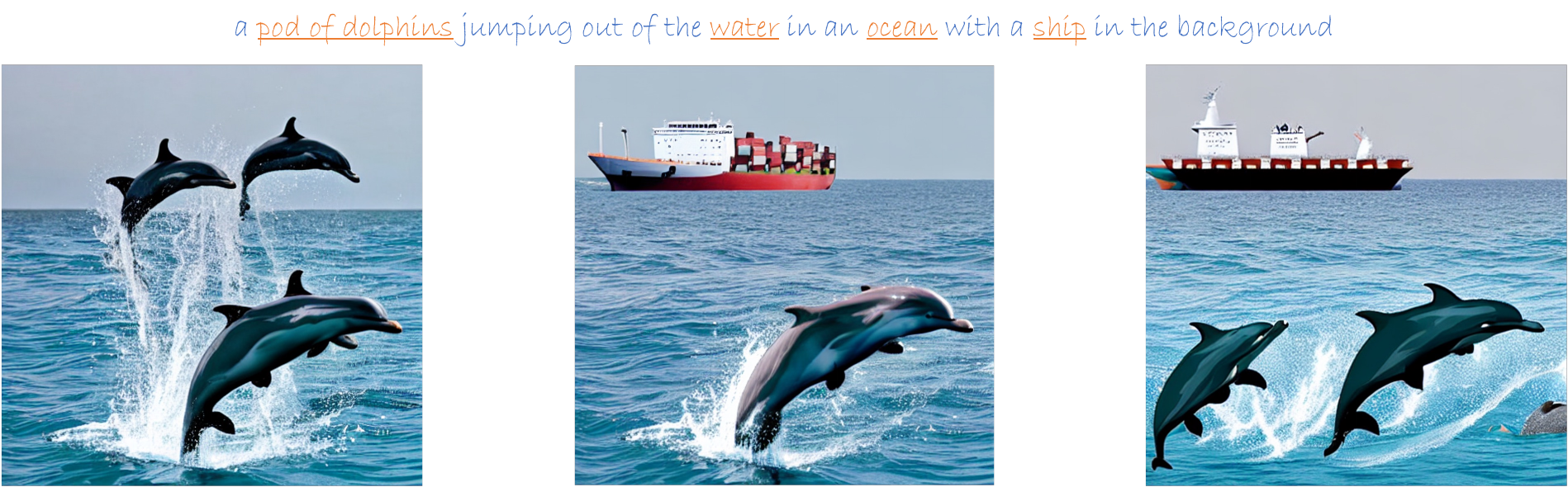}
     \end{subfigure}
\caption{{\emph{ Visualizing iterative refinement process for improving text-to-image alignment.}} 
}
\label{fig:iter-refinement-v2}
\end{center}
\vskip -0.3in
\end{figure}

\subsection{Additional Comparisons with Attend-and-Excite}
\label{sec:comp-attend-excite}

\begin{figure}[tbh]
\vskip 0.4in
\begin{center}
\centering
     \begin{subfigure}[b]{1.\columnwidth}
         \centering
         \includegraphics[width=1.\textwidth]{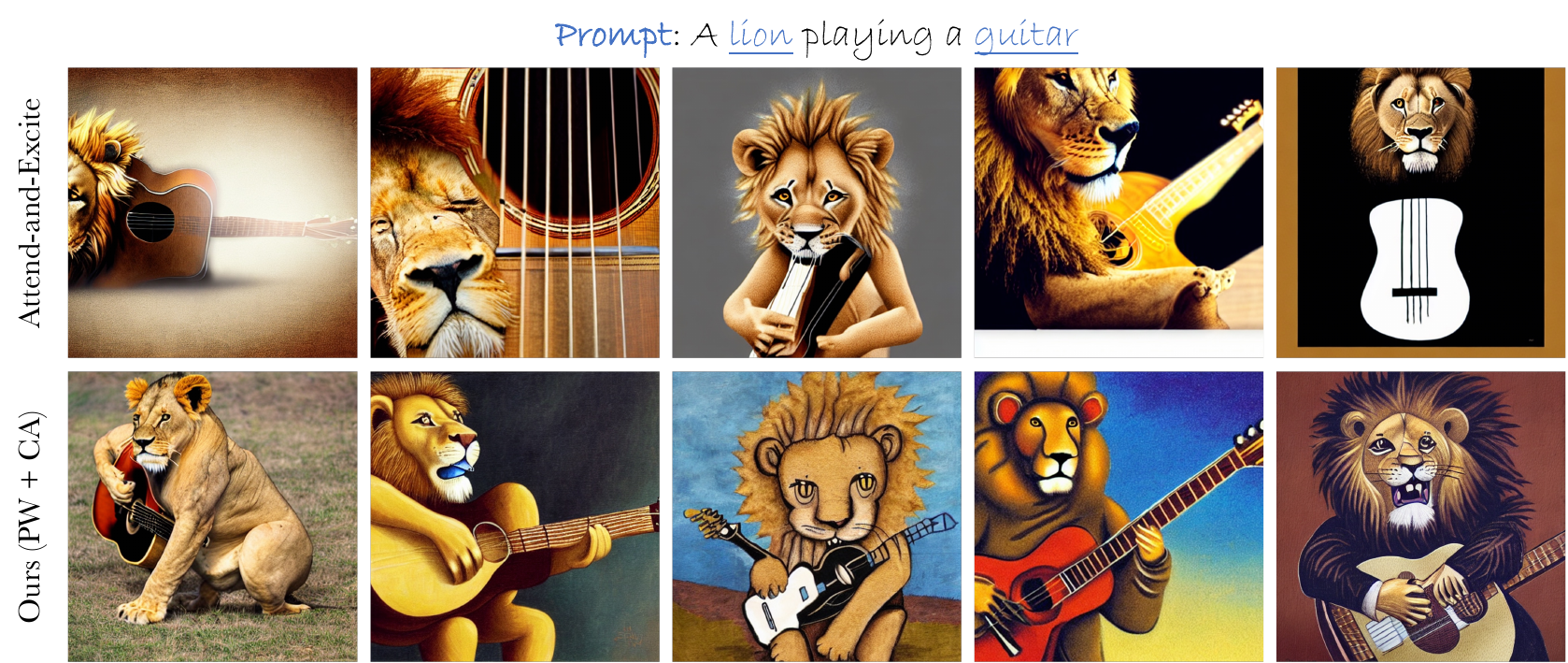}
     \end{subfigure}
\caption{{\emph{Additional Comparisons with Attend-and-Exite: Missing Relationship.}} Due to the pure object focused nature of Attend-and-Excite \cite{chefer2023attend}, it may result in images where all objects are present but the relationship between them is not accurately described. In contrast, we observe that the iterative refinement approach is able to better describe both presence and relationship between the objects.
}
\label{fig:attend-excite-p1}
\end{center}
\vskip 0.4in
\end{figure}

\begin{figure}[tbh]
\begin{center}
\centering
     \begin{subfigure}[b]{1.\columnwidth}
         \centering
         \includegraphics[width=1.\textwidth]{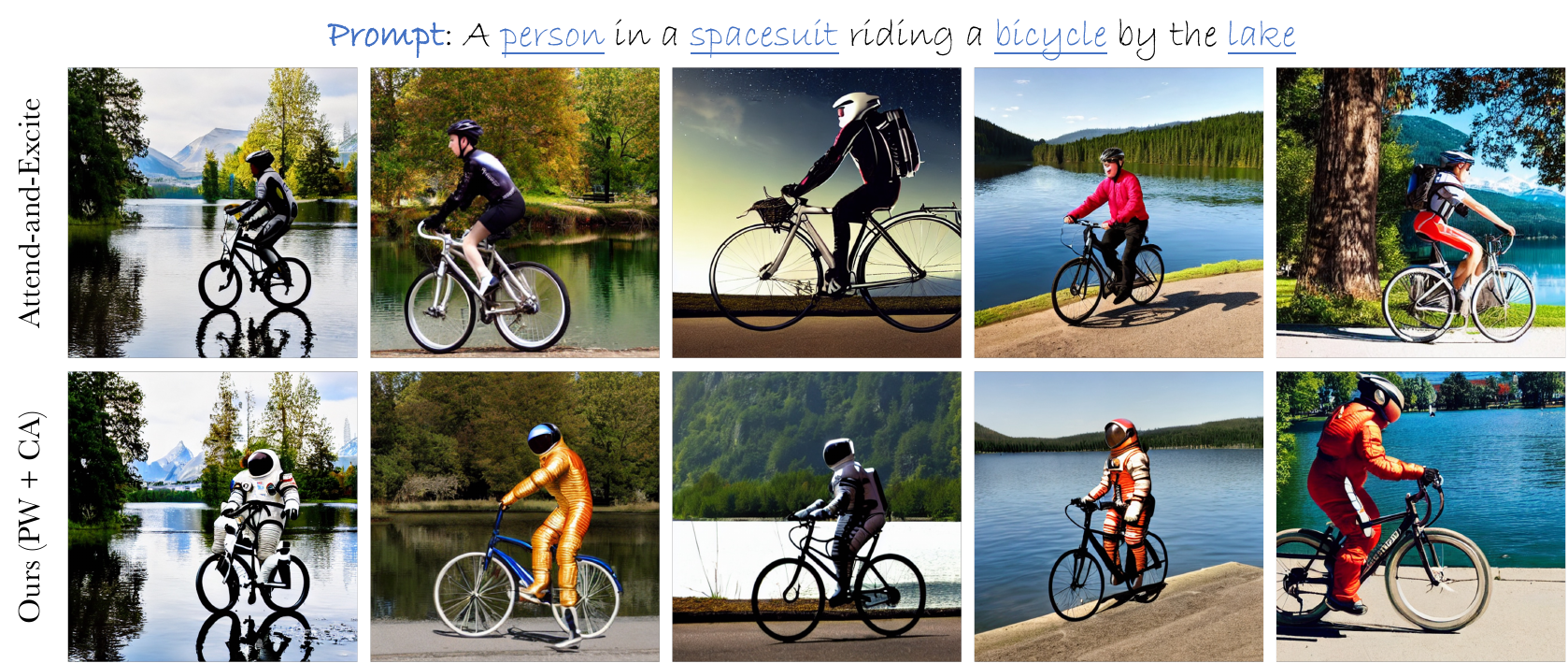}
     \end{subfigure}
\caption{{\emph{Additional Comparisons with Attend-and-Exite: Overlapping Entities.}} For images with overlapping  entities (\eg, \emph{\bline{person} and \bline{spacesuit}}), we observe that Attend-and-Excite \cite{chefer2023attend} typically spends most of gradient updates balancing between the overlapping entities, as both entities (\emph{\bline{person} and \bline{spacesuit}}) occupy the same cross-attention region. This can lead to outputs  where a) other important aspects (\eg, \emph{\bline{lake}} in Col-3) or b) one of the two entities (\eg, \emph{\bline{spacesuit}}) are ignored.
}
\label{fig:attend-excite-p1}
\end{center}
\end{figure}

\begin{figure}[tbh]
\begin{center}
\centering
     \begin{subfigure}[b]{1.\columnwidth}
         \centering
         \includegraphics[width=1.\textwidth]{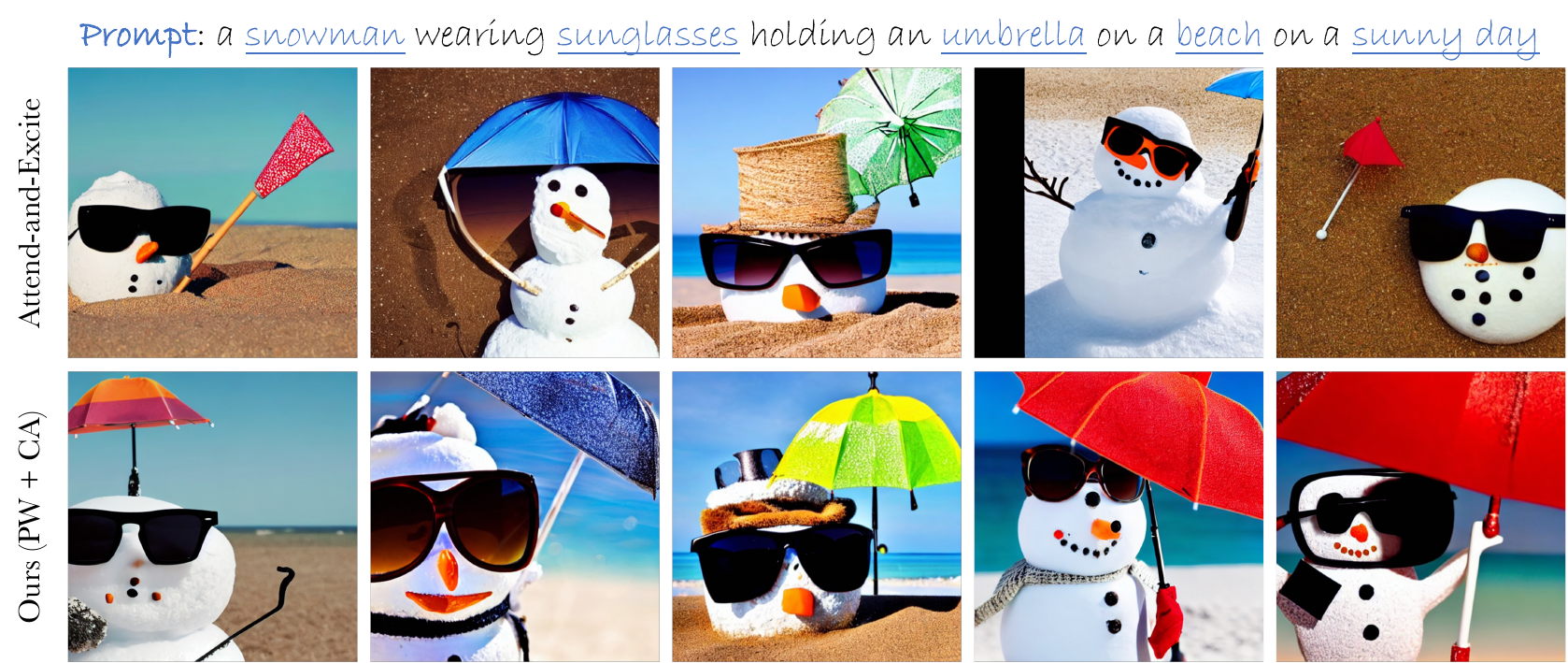}
     \end{subfigure}
\caption{{\emph{Additional Comparisons with Attend-and-Exite: Missing Objects.}} Since Attend-and-Excite \cite{chefer2023attend} is limited to applying the cross-attention update \emph{w.r.t} the least dominant subject, as the complexity of input prompt increases, it may miss some objects (\eg, umbrella, beach, sunny day) during the generation process. In contrast, the iterative nature of our approach allows it to keep refining the output image until a desirable threhold for the overall image-text alignment score is reached.
}
\label{fig:attend-excite-p1}
\end{center}
\vskip -0.1in
\end{figure}

\newpage
\subsection{Comparisons with Human-Feedback based Methods}
\label{sec:comp-hps}

\begin{figure}[b]
\vskip -0.1in
\begin{center}
    \centerline{\includegraphics[width=1.\linewidth]{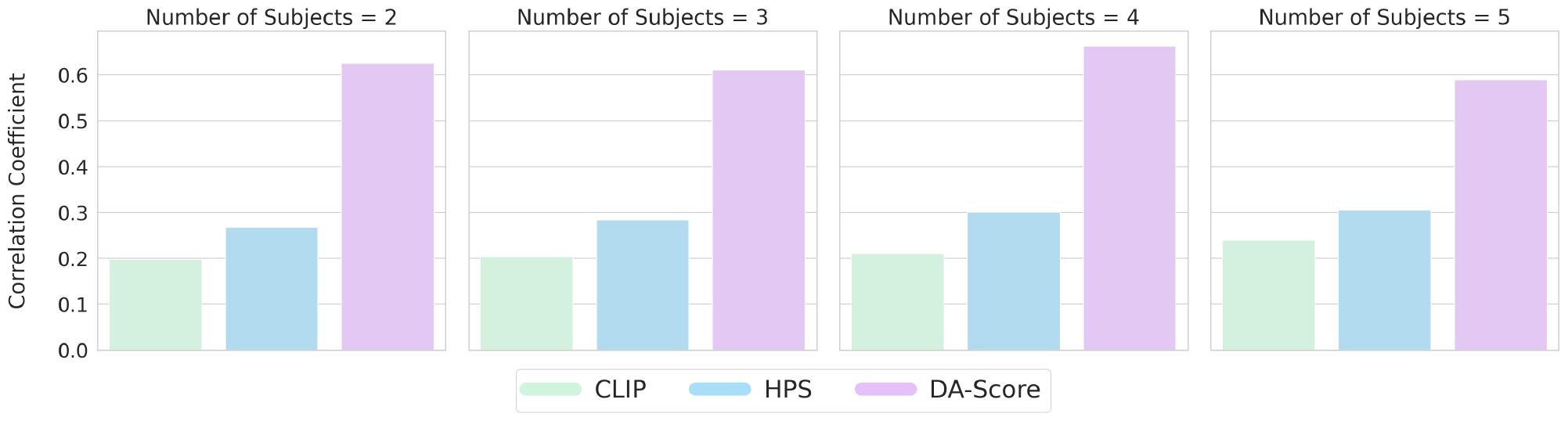}}
    \vskip -0.05in
    \caption{\emph{{Method comparisons w.r.t correlation with human ratings.}} We compare the correlation of HPS score (CLIP finetuned with human feedback) from Wu \etal \cite{wu2023better}. We observe that while the HPS score shows improved performance over CLIP \cite{radford2021learning}, the proposed DA-score shows better correlation with human ratings across varying number of subjects in input prompt.
    }
\label{fig:hps-corr}
\end{center}
\vskip -0.2in
\end{figure}

\begin{figure}[tbh]
\begin{center}
\centering
     \begin{subfigure}[b]{1.\columnwidth}
         \centering
         \includegraphics[width=1.\textwidth]{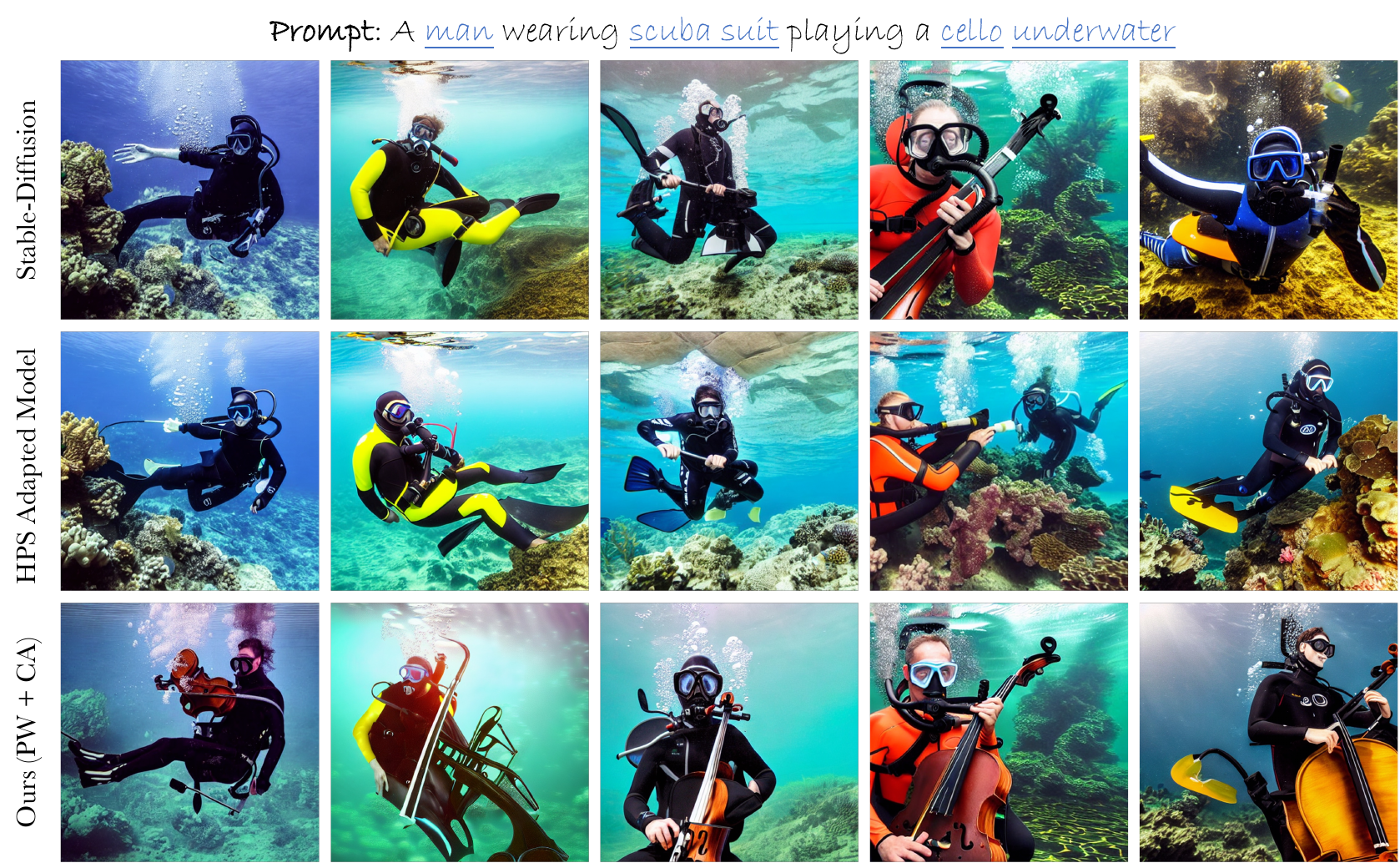}
     \end{subfigure}
     \vskip 0.05in
     \begin{subfigure}[b]{1.\columnwidth}
         \centering
         \includegraphics[width=1.\textwidth]{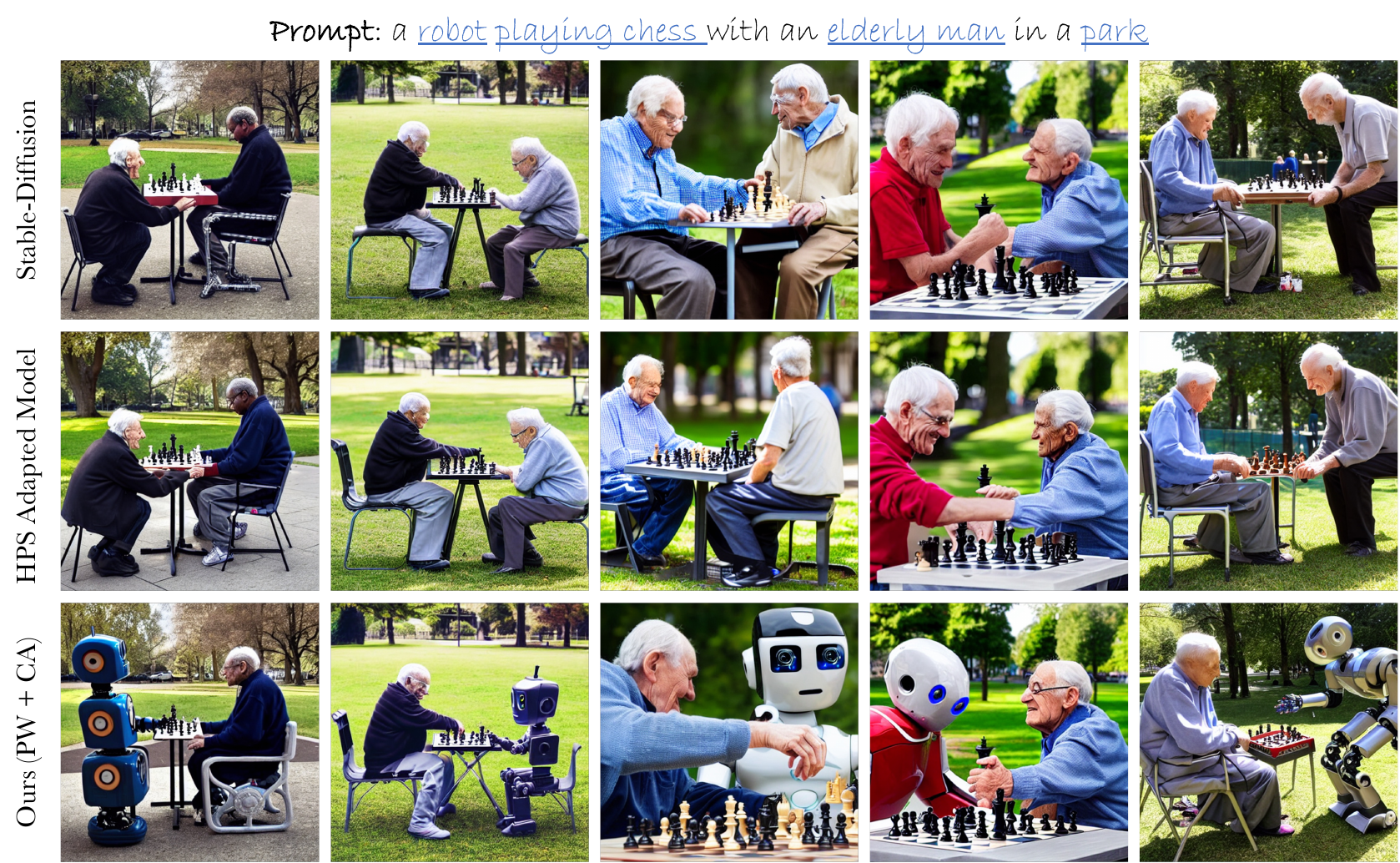}
     \end{subfigure}
\caption{{\emph{Comparing image alignment performance with human-feedback based models.}}  We find that while human-feedback based finetuned  diffusion model from \cite{wu2023better} improves the aesthetics (\eg lighting) of the output, it does not visually improve alignment with the input text for complex prompts.
}
\label{fig:hps-img-alignment-comp}
\end{center}
\end{figure}

Besides training-free methods, recent contemporary work \cite{lee2023aligning,wu2023better} has also explored the possibility of improving image-text alignment using human feedback to finetune existing latent diffusion models. For instance, Wu \etal \cite{wu2023better} recently release new versions of CLIP \cite{radford2021learning} and Stable Diffusion \cite{rombach2021highresolution} models which have been finetuned to better align with user preferences using human-feedback data. In this section, we compare the performance of our simple training-free approach with models released by \cite{wu2023better} in terms of both evaluation and improvement of fine-grain text-image matching.

\textbf{Results.} Fig.~\ref{fig:hps-corr} compares the correlation of 1) Original CLIP \cite{radford2021learning} model, 2) HPS scores (CLIP model finetuned to better align with human preference scores) by \cite{wu2023better}, and 3) DA-Scores predicted by our method on the Decomposable-Captions-4k dataset. We observe that while the HPS scores with the finetuned CLIP model, show significantly higher correlation with human annotations as opposed to original CLIP model, it still performs worse than the proposed DA-scores.

Similarly, Fig.~\ref{fig:hps-img-alignment-comp} compares the image outputs for 1) Original Stable Diffusion \cite{rombach2021highresolution} model, 2) HPS Adapted Stable Diffusion model from \cite{wu2023better} and 3) DA-score based iterative-refinement approach. We observe that while the HPS Adapted model improves the aesthetics of the generated model (\eg, improved lighting in Col-1,5 in example-1  Fig.~\ref{fig:hps-img-alignment-comp}), it does not improve the semantic alignment with content of the input prompt. In contrast, while our approach does not improve the aesthetics of the generated image, the output images show significantly higher alignment with the input prompt.

\textbf{Reason.} As shown above while human preference finetuned CLIP and Stable Diffusion models from \cite{wu2023better} show better performance in terms of aesthetics, they do not significantly improve visual alignment between generated images and input text as the complexity of the prompts increases. \emph{We believe that a major reason behind the same is the heavy data-driven nature of human-feedback based methods. That is, the generalization performance of the final finetuned model often relies heavily on the diversity and nature of the human-feedback dataset.}
In current works \cite{lee2023aligning,wu2023better}, human-feedback is typically collected by showing users $4-10$ images (for the same input prompt) generated by a pretrained Stable-Diffusion \cite{rombach2021highresolution} model, and then asking the users to select the best match.
However as shown in the main paper, as the complexity of the input prompt increases, the original Stable-Diffusion \cite{rombach2021highresolution} model shows a very low text-to-image aligment accuracy. As a result, the collected human-feedback data is often biased towards predicting more aesthically pleasing outputs, as opposed to outputs which improve fine-grain alignment with the content of the input prompt. 

In contrast, we propose a simple training-free approach which is able to generalize well to both simple and hard prompts, for both evaluation and improvement of text-to-image alignment. We also note that training-free methods such as ours or prior works \cite{chefer2023attend,liu2022compositional,feng2022training}, can in-turn help improve the performance of human-feedback based methods by providing a better quality dataset for determining both aesthetics and content alignment of the generated images.

\section{Implementation Details}
\label{sec:implementation-details}

In this section, we provide further details for the implementation of our approach as well as other baselines \cite{rombach2021highresolution,chefer2023attend,liu2022compositional,feng2022training} used while reporting results in the main paper. The full detailed implementation for both evaluation and improvement of text-to-image alignment is provided in Alg.~\ref{alg:vqa-score}, \ref{alg:vqa-iter}.

\textbf{Model Details.} Similar to \cite{chefer2023attend,feng2022training}, we use the official Stable Diffusion v1.4 model as the underlying pretrained text-to-image generation model while reporting results with all methods 
\cite{chefer2023attend,feng2022training,liu2022compositional} (including the iterative refinement approach proposed in the main paper). 
All results are reported at $512 \times 512$ resolution while using 50 inference steps during the reverse diffusion process. Unless otherwise specified, a fixed classifier-free guidance scale \cite{ho2022classifier} $\alpha_{cfg}=7.5$ is used for all experiments. 
By default, we use the pretrained BLIP-VQA \cite{li2022blip} model (BLIP model finetuned for visual question answering) for predicting the assertion-alignment scores while reporting all results.

\begin{figure}[t]
\vskip -0.15in
\begin{center}
\centering
     \begin{subfigure}[b]{1.\columnwidth}
         \centering
         \includegraphics[width=1.\textwidth]{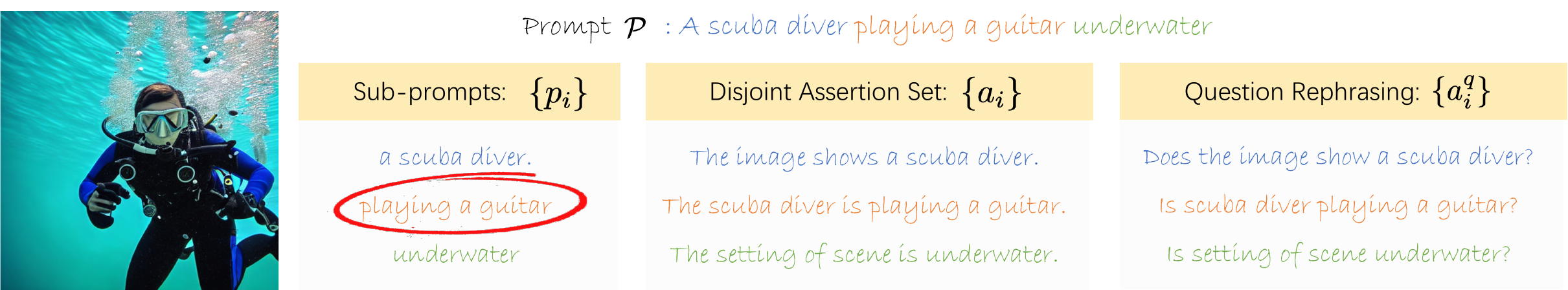}
     \end{subfigure}
     \vskip 0.05in
     \begin{subfigure}[b]{1.\columnwidth}
         \centering
         \includegraphics[width=1.\textwidth]{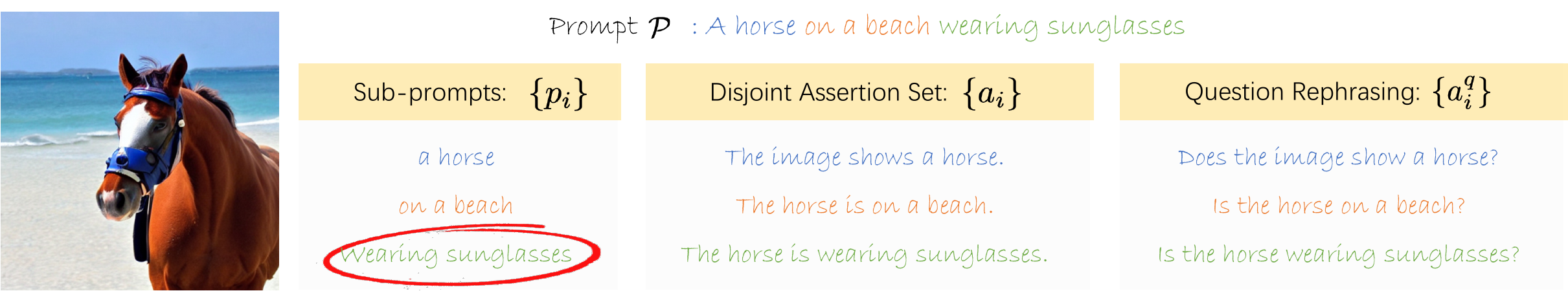}
     \end{subfigure}
     \vskip 0.05in
     \begin{subfigure}[b]{1.\columnwidth}
         \centering
         \includegraphics[width=1.\textwidth]{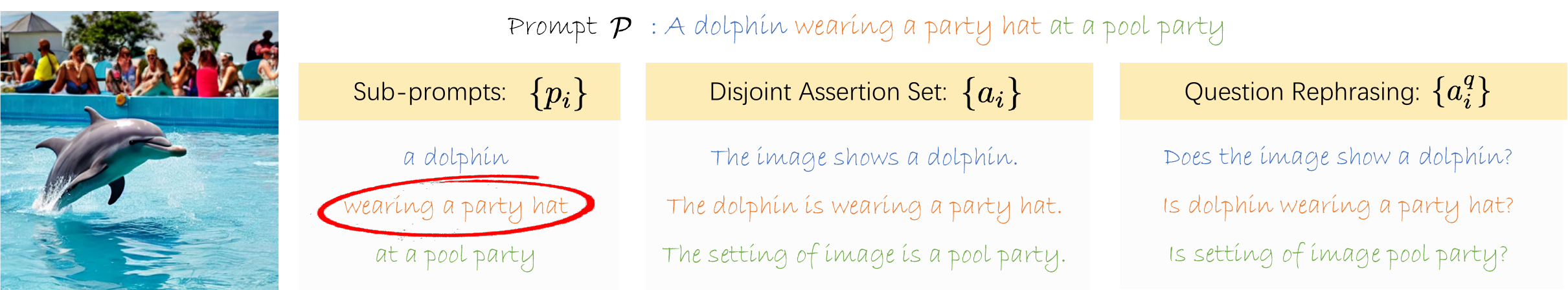}
     \end{subfigure}
\caption{{\emph{ Visualizing the outputs of prompt decomposition.}} 
By dividing a complex prompt $\mathcal{P}$ into a set of disjoint assertions $a_i$, we are able to identify the sub-prompts  $p_i$ (circled) which are not being expressed in the image output using VQA, and thereby address them using iterative refinement.}
\label{fig:prompt-decomposition}
\end{center}
\vskip -0.1in
\end{figure}

\textbf{Prompt Decomposition.} Similar to our approach, prior works on improving image-text alignment often rely on human-user inputs for expressing contents of the input prompt into its simpler constituents. For instance, Feng \etal \cite{liu2022compositional} require the user to describe the prompt as a conjunction / disjunction of simpler statements. Chefer \etal \cite{chefer2023attend} require the user to provide a set of entities / subjects in the prompt, over which the cross-attention optimization should be performed. Similarly, in order to evaluate the \emph{Decompositional Alignement Scores}, our approach relies on decomposing the input caption into a set of disjoint assertions (with their rephrasing as a question). 

Instead of relying of human inputs as in prior works, we leverage the in-context learning capability of large-language models (LLMs) \cite{liu2023summary,touvron2023llama} for obtaining such decompositions in an autonomous manner.
This allows us to perform large-scale quantitative evaluations across different methods in a robust manner.
In particular, for all methods \cite{chefer2023attend, liu2022compositional} and ours, we first collect a set of 4-5 human generated examples describing the desired outputs (\eg, main subjects for \cite{chefer2023attend}) for the input prompts. This example-set is then used as an in-context dataset which is then fed to the \emph{`gpt-4'} \cite{liu2023summary} model to generate the desired prompt decompositions on prompts across the \emph{Decomposable-Captions-4k} dataset.
Fig.~\ref{fig:prompt-decomposition} provides an overview of the different prompt decomposition outputs for our approach.

Note that while it is possible to explore relatively simpler methods \eg, extracting noun phrases for subject extraction in \cite{chefer2023attend}, it can lead to errors as the complexity of the input prompt increases. For instance, for the input \emph{prompt: ``a \bline{penguin} wearing a \bline{bowtie} with a bright \bline{sun} in the background''}, the extracted noun phrases will also include the word \emph{``background''}. The use of a LLM-based in-context framework allows us to avoid such errors.
In future, the proposed approach can be extended to allow for much lightweight decomposition by using a low-rank finetuning \cite{hu2021lora} for adaptation of recently released instruction-following models \cite{alpaca}. However, since the same is not the main focus of our work, we leave it as a directive for future research.

\textbf{Hyperparameters and Overall Algorithm.} The proposed iterative refinement approach uses a maximum of $K=5$ iterations for the refinement process. The iterative process is terminated early if a  threshold of $0.8$ for the overall image alignment score $\Omega (\mathcal{I}_k,\mathcal{P})$ is obtained. The iterative refinement weights are initialized as $w_i = 1 \forall i$  for prompt weighting (PW), and, $\gamma_i = 0 \forall i$ for cross-attention (CA) updates. An increment $\Delta$ of $0.1$ and $1.0$ is used for updating the assertion weights for prompt-weighting (PW) and cross-attention (CA) update methods respectively. Furthermore, to reduce the inference time for each iteration as compared to \cite{chefer2023attend}, cross-attention updates are only applied for first 20 steps of the reverse diffusion process. Furthermore, the use of iterative cross attention updates is also discarded. 
The image generation output $\mathcal{I}^\star$ at the end of the refinement process is computed as,
\begin{align}
    \mathcal{I}^\star = \text{argmax}_{\mathcal{I}_k} \Omega(\mathcal{I}_k,\mathcal{P}).
\end{align}
Please refer Alg.~\ref{alg:vqa-score}, \ref{alg:vqa-iter} for the full detailed implementation (with hyperparameters) for our approach.

\begin{algorithm}[tbh]
	\caption{DA-Score: Evaluating Text-to-Image Alignment}
    \textbf{Input}: Text prompt $\mathcal{P}$, generated image $\mathcal{I}$.\\
    \textbf{Output}: Text-to-Image Alignment Score between $\mathcal{P}, \mathcal{I}$\\
    \textbf{Require}: Large-language model $\mathcal{M}$, VQA-model $\mathcal{V}$, exempler dataset $\mathcal{D}$, task description $\mathcal{T}$, softmax-temperature $\tau = 0.9$
    \begin{algorithmic}[1]
            \State {$\triangleright$ \textsc{\bline{Prompt Decomposition}}}
            \State $\mathbf{x} = \{x_0,x_1, \dots x_n\} = \mathcal{M}(\mathbf{x} \mid \mathcal{P},\mathcal{D}_{exempler},\mathcal{T}), \quad \textit{where} \ \ x_i = \{a_i, p_i, a^q_i\}$;
            \\
            \State {$\triangleright$ \textsc{\bline{Compute Assertion Alignment Scores using VQA}}}
            \State $u_i(\mathcal{I}, a_i) = \frac{\exp{(\alpha_i/\tau)}}{\exp{(\alpha_i/\tau)} + \exp{(\beta_i/\tau)}}, \ \textit{where} \quad \alpha_i = \mathcal{V} (\textit{`yes'} \mid \mathcal{I}, a^q_i), \quad  \beta_i = \mathcal{V} (\textit{`no'} \mid \mathcal{I}, a^q_i)$
            \\
            \State {$\triangleright$ \textsc{\bline{Overall Image-Text Alignment Score}}}
            \State $\Omega(\mathcal{I},\mathcal{P}) =  {\sum_i \lambda_i(\mathcal{P},a_i) \ u_i(\mathcal{I}, a_i)} / {\sum_i \lambda_i(\mathcal{P},a_i)},$
            \\
            \State \Return $\Omega(\mathcal{I},\mathcal{P})$.
	\end{algorithmic}
	\label{alg:vqa-score}
\end{algorithm}

\begin{algorithm}[tbh]
	\caption{Iterative Refinement: Improving Text-to-Image Alignment}
    \textbf{Input}: Text prompt $\mathcal{P}$, subprompts $p_i$, disjoint assertions in question format $a^q_i$.\\
    \textbf{Output}: Image generation output $\mathcal{I}^\star$ conditioned on $\mathcal{P}$\\
    \textbf{Require}: Pretrained diffusion model $\mathcal{D}$, VQA-model $\mathcal{V}$, prompt-weighting function $\mathcal{W}$ \\
    \textbf{Hyperparameters}: max number of iterations $K=5$, alignment threshold $\Omega_{max}=0.8$, weight increments $\Delta_{w}=0.1$, $\Delta_{\gamma}=1.0$, step-size $\alpha=10$, number of cross-attention update steps $t_0 = 20$.
    \begin{algorithmic}[1]
        \State {$\triangleright$ \textsc{\bline{Initialize Assertion Weights}}}
        \State Initialize \ $ w^{0}_i = 1 \ \forall i$; \Comment{for prompt-weighting}
        \State Initialize \ $ \gamma^{0}_i = 0 \ \forall i$; \Comment{for cross-attention updates}
        \\
            \State {$\triangleright$ \textsc{\bline{Iterative Refinement}}}
            \For{$0 \leq k < K$}
                \\
                \State {$\triangleright$ \textsc{\bline{prompt weighting}}}
                \State $y_{prompt} = \mathcal{W}(\mathcal{P},\{\text{CLIP}(p_i),w^k_i\}_{i=1}^n))$; 
                \\
                \State {$\triangleright$ \textsc{\bline{Parameterized Reverse Diffusion}}}
                \State Sample \ $z_{T} \sim \mathcal{N}(0,\bm{I})$;
                \For{$0 < t \leq T$}
                    \State $\_, \mathcal{A}_i^t = \mathcal{D}(z_t,y_{prompt},t)$;\Comment{compute cross-attention maps}
                    \If{$t \geq T - t_0$}
                    \State {$\triangleright$ \textsc{\bline{Weighted Cross-Attention Updates}}}
                     \State $\mathcal{L}(z_t, \{\gamma^k_i\}_{i=1}^n) = \sum_i \gamma^k_i (1- \text{max} \  G(\mathcal{A}^t_i))$;
                     \State $z_t = z_t - \alpha \nabla_{z_t} \mathcal{L}(z_t, \{\gamma^k_i\}_{i=1}^n))$;
                    \EndIf
                    \State $z_{t-1} = \textsc{ReverseDiff}(z_{t},t \rightarrow t-1 \mid y_{prompt})$;
                \EndFor
                \\
                \State {$\triangleright$ \textsc{\bline{Get Decompositional-Alignment Scores}}}
                \State $\mathcal{I}_k = x_0 = \textsc{Decoder}(z_0)$;
                \State $\Omega(\mathcal{I}_k,\mathcal{P}), \{u_i(\mathcal{I}_k,\mathcal{P})\}_{i}^n = \textsc{DA-Score}(\mathcal{I}_k, \{a^q_i\}_i)$;
                \\
                \State {$\triangleright$ \textsc{\bline{Finish If Output is Good Enough}}}
                \If{$\Omega(\mathcal{I}_k,\mathcal{P}) \geq \Omega_{max}$}
                    \State \Return $\mathcal{I}^\star = \mathcal{I}_k$.
                \EndIf
                \\
                \State {$\triangleright$ \textsc{\bline{Update Assertion Weights}}}
                \State $w_i^{k+1} = 
                \begin{cases}
                w_i^k + \Delta_w, \ \text{if} \quad  i = \text{argmin}_{l}  \ u_l(\mathcal{I}_k,\mathcal{P})\\
                w_i^k \quad \text{otherwise}
                \end{cases}$ \Comment{for prompt-weighting}
                \State $\gamma_i^{k+1} = 
                \begin{cases}
                \gamma_i^k + \Delta_\gamma, \ \text{if} \quad  i = \text{argmin}_{l}  \ u_l(\mathcal{I}_k,\mathcal{P})\\
                \gamma_i^k \quad \text{otherwise}
                \end{cases}$ \Comment{for cross-attention updates}
            \EndFor
            \\
            \State {$\triangleright$ \textsc{\bline{Return Best Output}}}
            \State \Return $\mathcal{I}^\star = \text{argmax}_{\mathcal{I}_k} \Omega(\mathcal{I}_k,\mathcal{P})$.
	\end{algorithmic}
	\label{alg:vqa-iter}
\end{algorithm}

\section{Decomposable Captions 4K Dataset}
\label{sec:decomposable-captions}

\begin{figure}[h!]
\begin{center}
\centering
     \begin{subfigure}[b]{0.99\columnwidth}
         \centering
         \includegraphics[width=1.\textwidth]{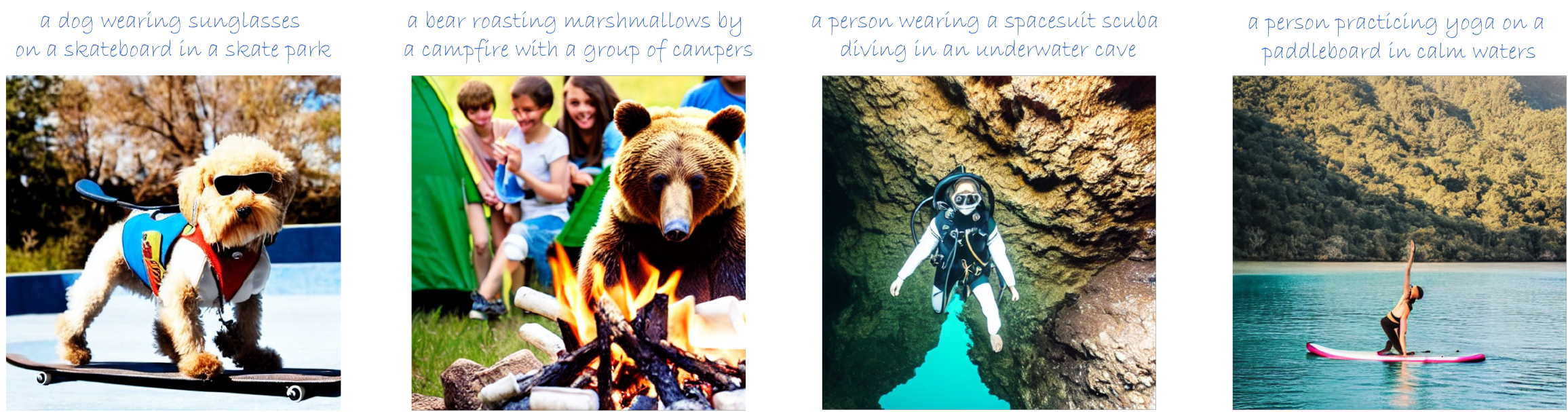}
         \vskip -0.02in
         \caption{\emph{Example Samples with User Rating: 5 (Image is an accurate match for the prompt)}}
     \end{subfigure}
     \vskip 0.05in
     \begin{subfigure}[b]{0.99\columnwidth}
         \centering
         \includegraphics[width=1.\textwidth]{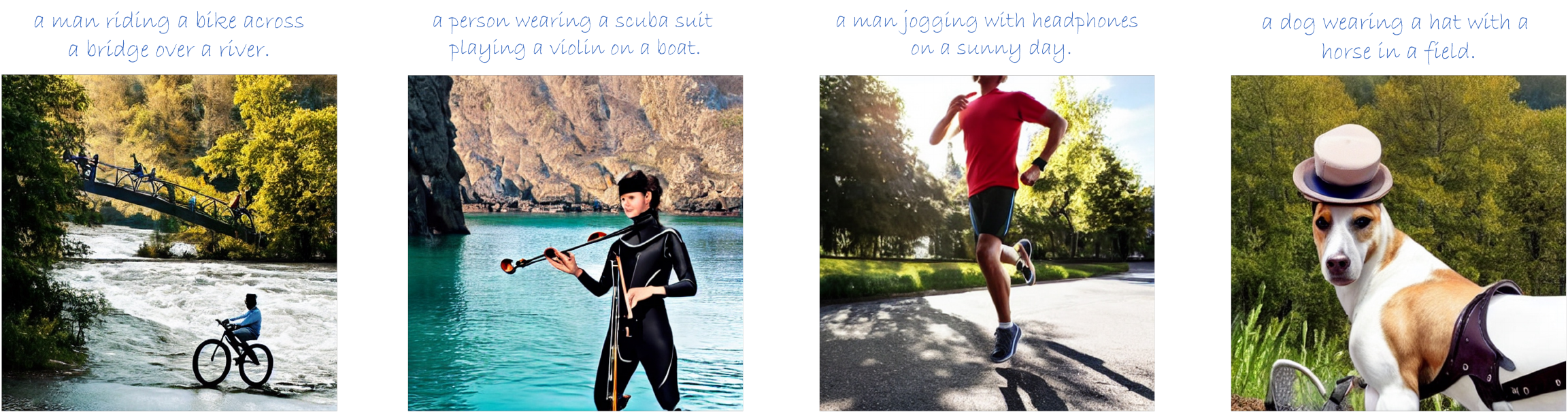}
         \vskip -0.02in
         \caption{\emph{Example Samples with User Rating: 4 (Image is a good match for prompt with minor mistakes)}}
     \end{subfigure}
     \vskip 0.05in
     \begin{subfigure}[b]{0.99\columnwidth}
         \centering
         \includegraphics[width=1.\textwidth]{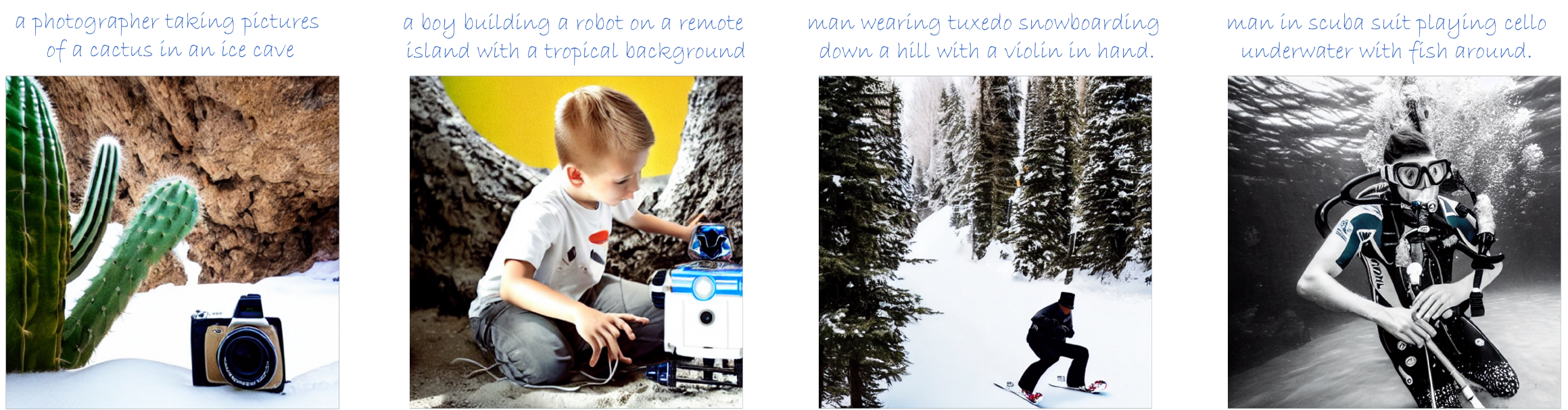}
         \vskip -0.02in
         \caption{\emph{Example Samples with User Rating: 3 (Image seems like a 50-50 match for the prompt)}}
     \end{subfigure}
     \vskip 0.05in
     \begin{subfigure}[b]{0.99\columnwidth}
         \centering
         \includegraphics[width=1.\textwidth]{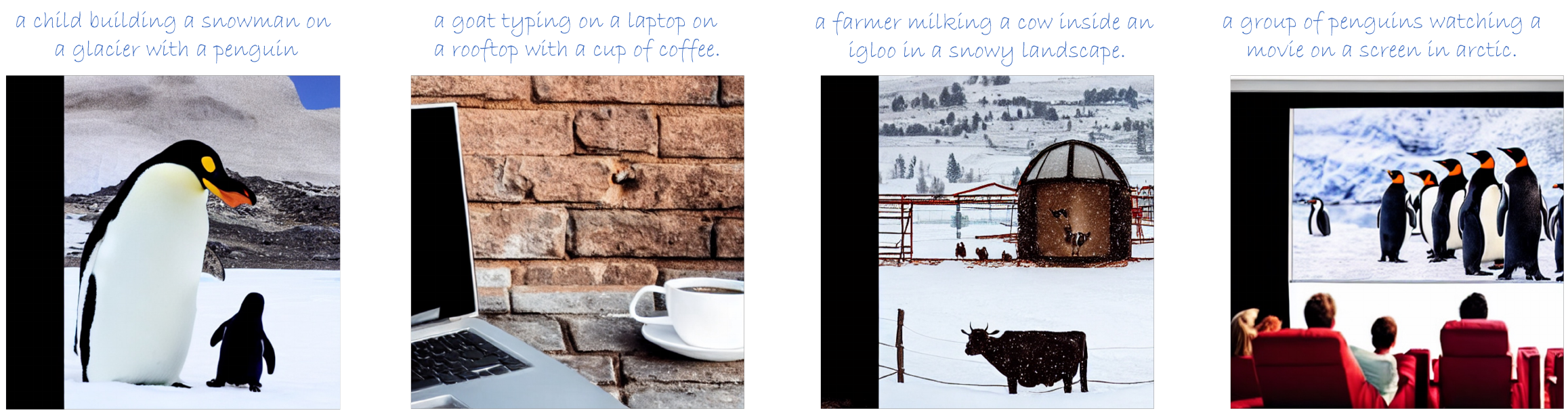}
         \vskip -0.02in
         \caption{\emph{Example Samples with User Rating: 2 (Image captures minor aspects about the input prompt)}}
     \end{subfigure}
     \vskip 0.05in
     \begin{subfigure}[b]{0.99\columnwidth}
         \centering
         \includegraphics[width=1.\textwidth]{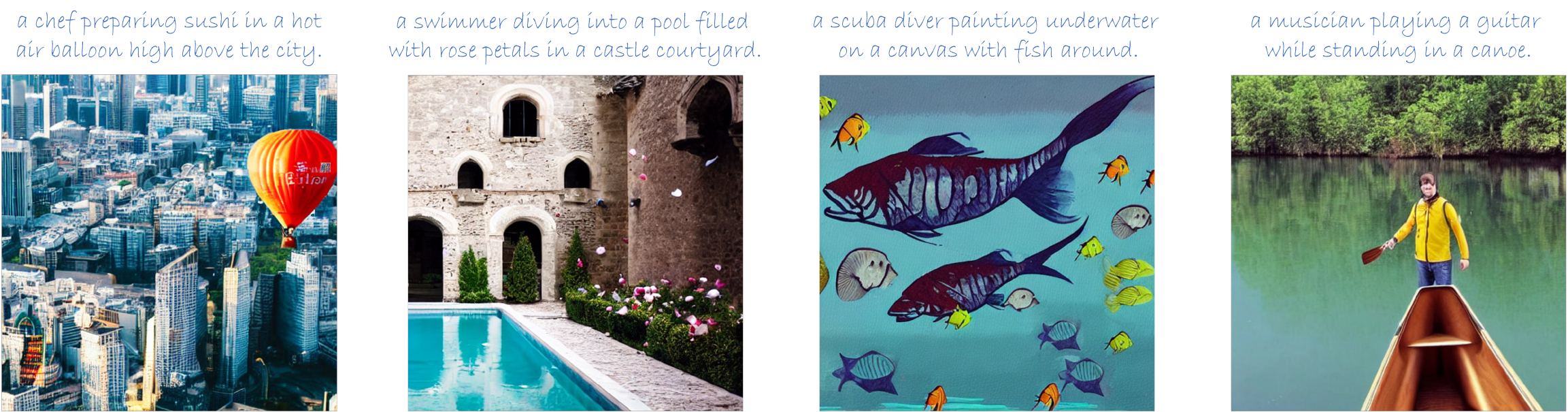}
         \vskip -0.02in
         \caption{\emph{Example Samples with User Rating: 1 (Image seems semantically irrelevant to the prompt)}}
     \end{subfigure}
\caption{{\emph{Visualizing samples with human annotations from the Decomposable-Captions-4k dataset.}} 
}
\label{fig:decomposable-captions-dataset-4k-rating}
\end{center}
\vskip -0.1in
\end{figure}

\textbf{Overview.} Since there are no openly available datasets addressing semantic challenges in text-based image generation with human annotations, we introduce a new benchmark dataset Decomposable-Captions-4k for method comparison. The dataset consists an overall of 24960 human annotations on images generated using all methods \cite{rombach2021highresolution,liu2022compositional,chefer2023attend} (including ours) across a diverse set of 4160 input prompts.  Each image is a given rating between 1 and 5 (where 1 represents that \textit{`image is semantically irrelevant to the prompt'} and 5 represents that \textit{`image is an accurate match for the prompt'}). Fig.~\ref{fig:decomposable-captions-dataset-4k-rating} provides an overview of some user annotations for image-prompt pairs from the curated dataset.

\textbf{Collecting Diverse Prompts of Varying Complexity.}
Furthermore, unlike prior works \cite{chefer2023attend} which predominantly analyse the performance on relatively simple prompts with two subjects (\eg, object a and object b), we construct a systematically diverse pool of input prompts for better understanding text-to-image alignment across varying complexities in the text prompt. In particular, the prompts for the dataset are designed to encapsulate two axis of complexity: \emph{number of subjects} and \emph{realism}. The number of subjects refers to the number of main objects described in the input prompt and varies from 2 (\eg, \emph{a \bline{cat} with a \bline{ball}}) to 5 (\eg, \emph{a \bline{woman} walking her \bline{dog} on a \bline{leash} by the \bline{beach} during \bline{sunset}}). Similarly, the \emph{realism} of a prompt is defined as the degree to which different concepts naturally co-occur together and varies as \emph{easy}, \emph{medium}, \emph{hard} and  \emph{very hard}. \emph{easy} typically refers to prompts where concepts are naturally co-occurring together (\eg, \emph{a \bline{dog} in a \bline{park}}) while \emph{very hard} refers to prompts where concept combination is very rare (\eg, \emph{a \bline{dog} playing a \bline{piano}}).

\begin{figure}[t]
\begin{center}
\centering
     \begin{subfigure}[b]{1.\columnwidth}
         \centering
         \includegraphics[width=1.\textwidth]{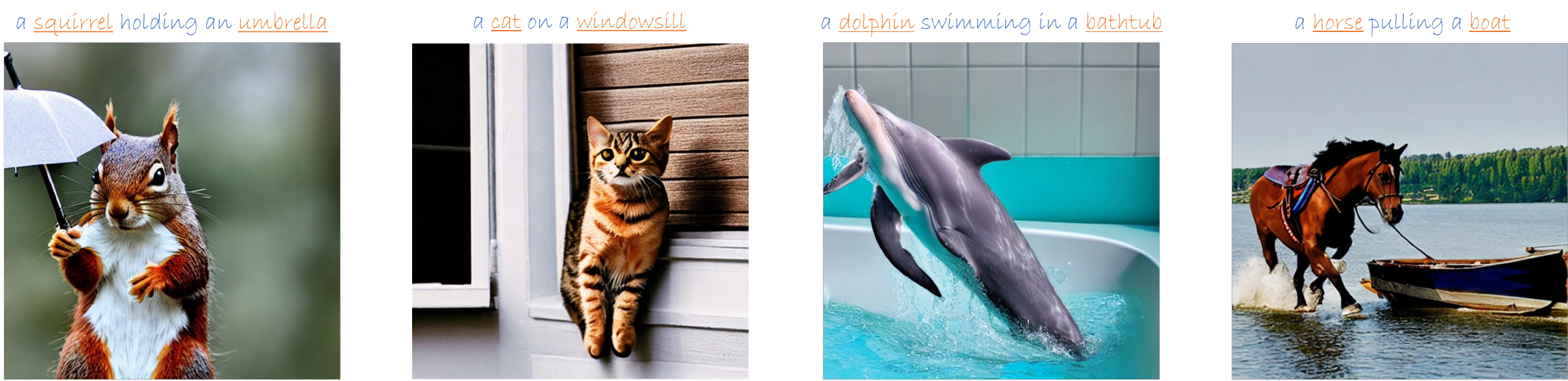}
         \vskip -0.02in
         \caption{\emph{Example sample outputs with our method for number of subjects: 2}}
     \end{subfigure}
     \vskip 0.05in
     \begin{subfigure}[b]{1.\columnwidth}
         \centering
         \includegraphics[width=1.\textwidth]{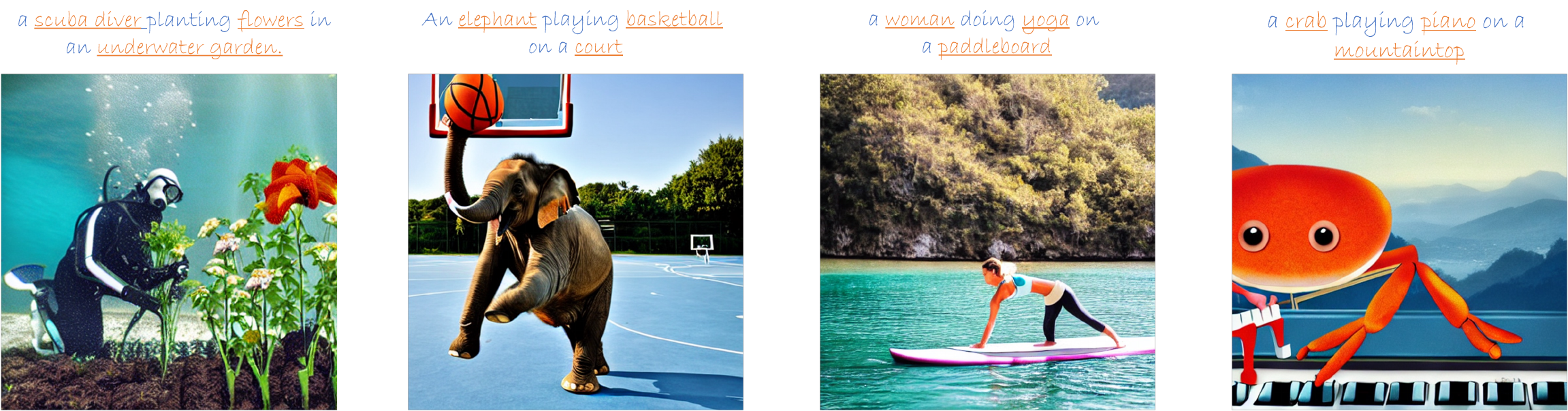}
         \vskip -0.02in
         \caption{\emph{Example sample outputs with our method for number of subjects: 3}}
     \end{subfigure}
     \vskip 0.05in
     \begin{subfigure}[b]{1.\columnwidth}
         \centering
         \includegraphics[width=1.\textwidth]{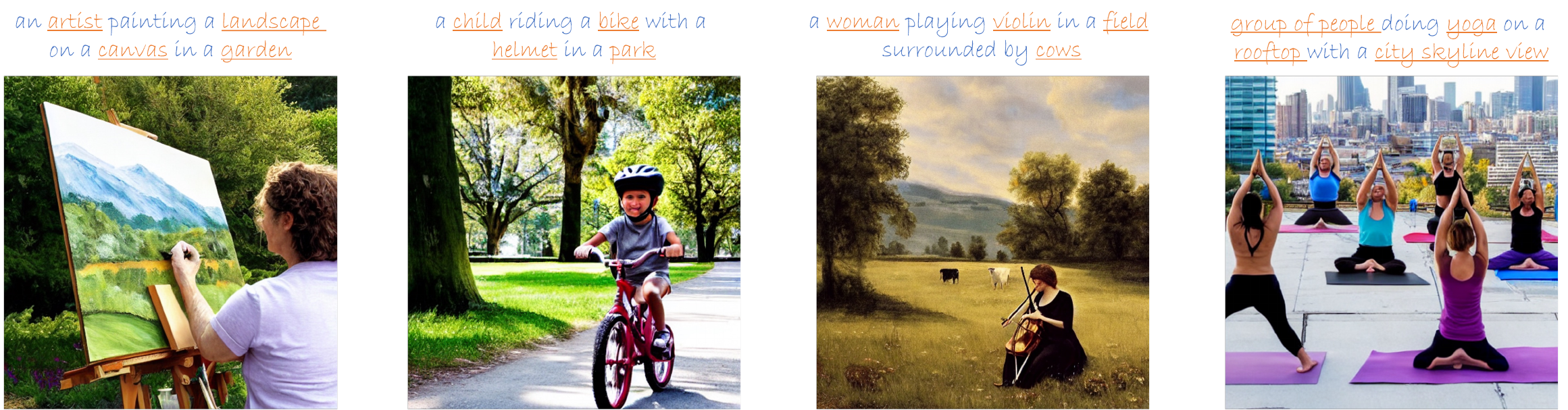}
         \vskip -0.02in
         \caption{\emph{Example sample outputs with our method for number of subjects: 4}}
     \end{subfigure}
     \vskip 0.05in
     \begin{subfigure}[b]{1.\columnwidth}
         \centering
         \includegraphics[width=1.\textwidth]{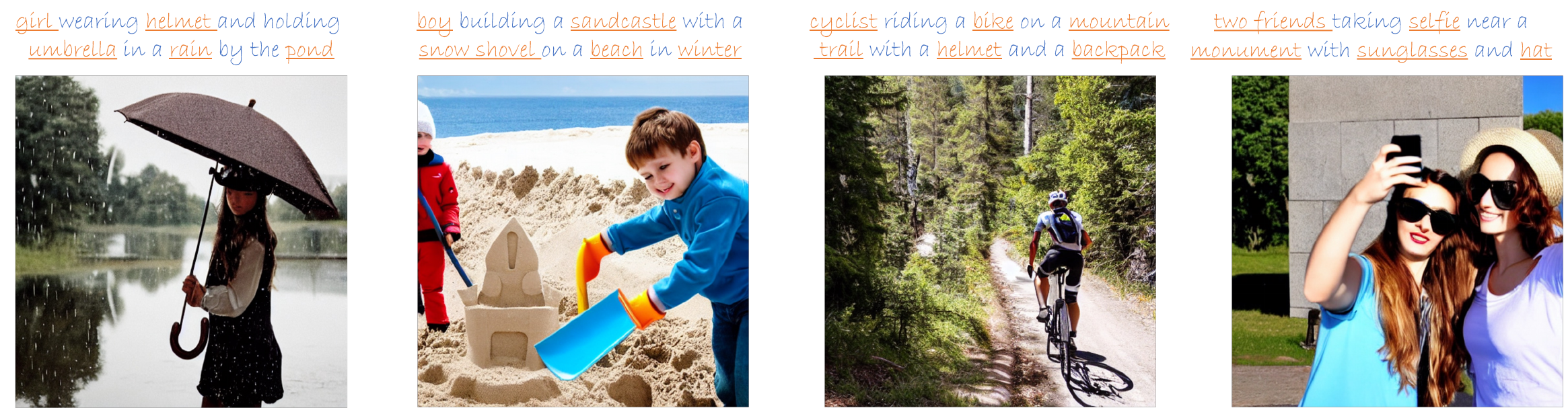}
         \vskip -0.02in
         \caption{\emph{Example sample outputs with our method for number of subjects: 5}}
     \end{subfigure}
\caption{{\emph{Visualizing variation in number of subjects in prompts from the Decomposable-Captions-4k dataset. All images are generated using the proposed iterative refinement approach.}}
}
\label{fig:decomposable-captions-dataset-4k-num-subjects}
\end{center}
\vskip -0.1in
\end{figure}

\begin{figure}[t]
\begin{center}
\centering
     \begin{subfigure}[b]{1.\columnwidth}
         \centering
         \includegraphics[width=1.\textwidth]{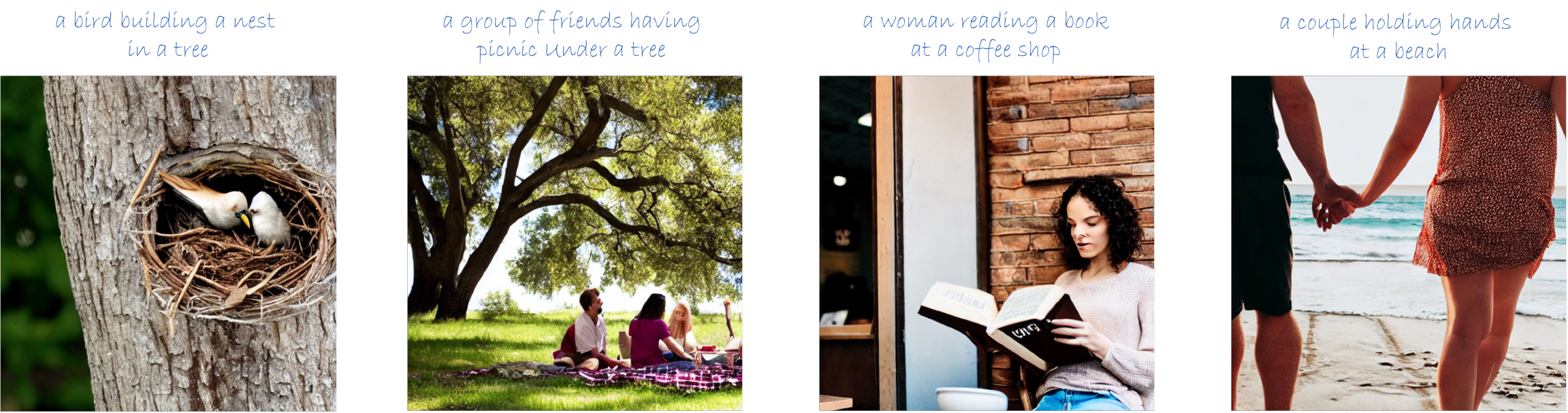}
         \vskip -0.02in
         \caption{\emph{Example sample outputs with our method for realism difficulty: easy}}
     \end{subfigure}
     \vskip 0.05in
     \begin{subfigure}[b]{1.\columnwidth}
         \centering
         \includegraphics[width=1.\textwidth]{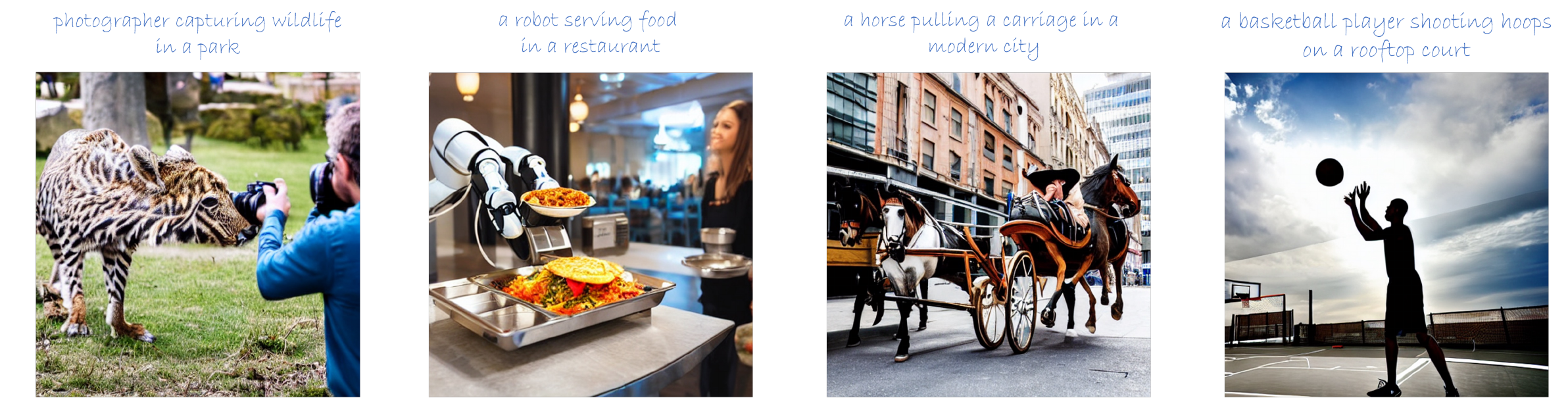}
         \vskip -0.02in
         \caption{\emph{Example sample outputs with our method for realism difficulty: medium}}
     \end{subfigure}
     \vskip 0.05in
     \begin{subfigure}[b]{1.\columnwidth}
         \centering
         \includegraphics[width=1.\textwidth]{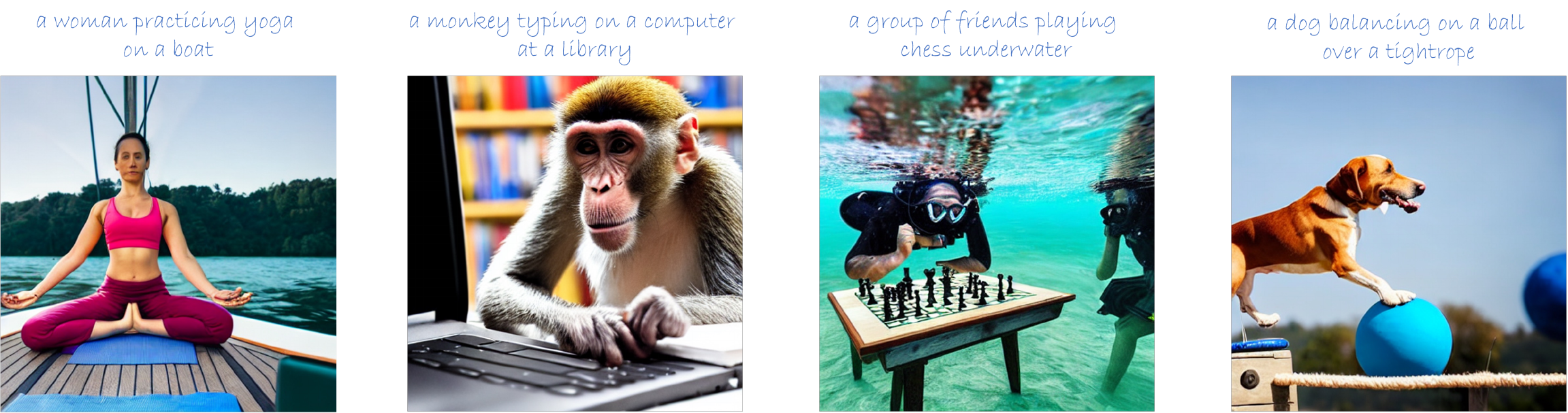}
         \vskip -0.02in
         \caption{\emph{Example sample outputs with our method for realism difficulty: hard}}
     \end{subfigure}
     \vskip 0.05in
     \begin{subfigure}[b]{1.\columnwidth}
         \centering
         \includegraphics[width=1.\textwidth]{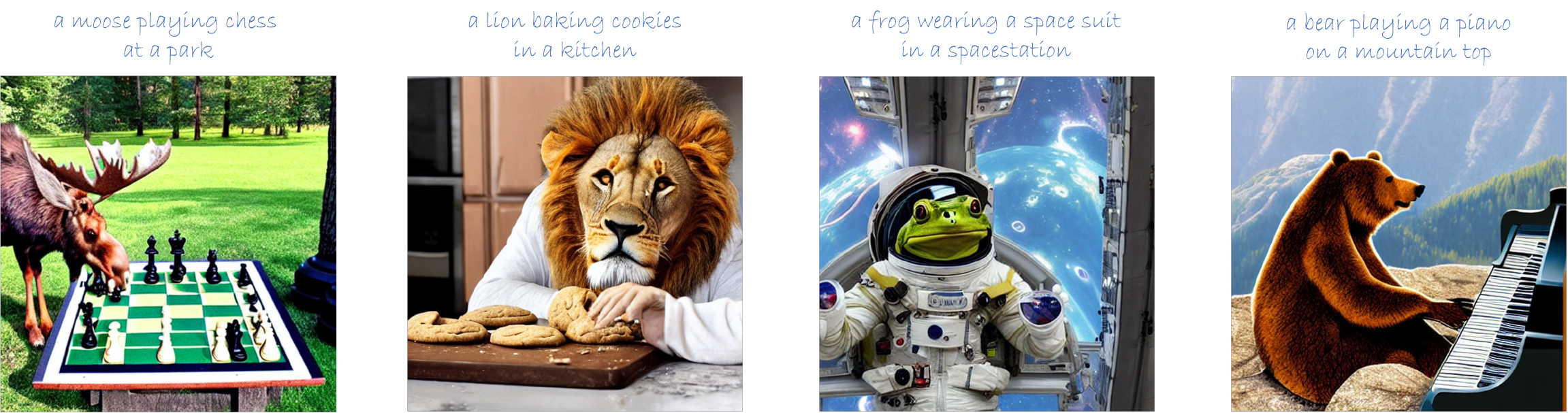}
         \vskip -0.02in
         \caption{\emph{Example sample outputs with our method for realism difficulty: very hard}}
     \end{subfigure}
\caption{{\emph{Visualizing variation in realism difficulty from the Decomposable-Captions-4k dataset. All prompts have been sampled using number of subjects = 3 subset of the overall dataset.}} 
}
\label{fig:decomposable-captions-dataset-4k-realism}
\end{center}
\vskip -0.1in
\end{figure}

\textbf{Prompt Generation.} A key part of the \emph{Decomposable-Captions-4k} dataset is to collect a set of diverse input prompts of varying complexity which would allow for a much more comprehensive evaluation across different methods. Moreover unlike prompts found in typical large-scale image-text datasets \cite{schuhmann2021laion,schuhmann2022laion}, the generated prompts should be imaginative and be able to describe novel and often non-realistic combinations of different concepts (\eg \emph{a \bline{lion} playing a \bline{piano}}). 

To this end, we leverage the diverse language modelling capabilities of large-scale large language models (LLM) \cite{liu2023summary,touvron2023llama} in order to generate novel prompts of varying complexity and realism. In particular, given a desired number of subjects $N$ (between 2 and 5), we first use the \emph{GPT-4} model \cite{liu2023summary} API with 8K context length to come up with an initial random subject \eg, a \bline{dog}. The model is then instructed to conditionally generate a second subject (\eg, \bline{sunglasses}) which is then combined with first subject to generate the sub-prompt \emph{``a \bline{dog} wearing \bline{sunglasses}''}. This process is continued until a complete prompt with a desired number of subjects is obtained (\eg, \emph{a \bline{dog} wearing \bline{sunglasses} on a \bline{skateboard} in a \bline{park}}). At the end of generation process, prompts which are grammatically inaccurate are filtered and removed. An overview of example prompts with variable number of subjects (along with corresponding image generation outputs with our approach) is provided in Fig.~\ref{fig:decomposable-captions-dataset-4k-num-subjects}.

Furthermore, in order to generate prompts of varying level of \emph{realism difficulty}, we generate prompts in batches of 4. In particular, during the sequential generation process (described above) the model is prompted to generate prompts of increasing level of \emph{realism difficulty} by asking it generate additional subjects whose combination in a sentence is increasingly more rare. For instance, for \emph{realism difficulty: easy}, the model is tasked to generate additional subjects which typically co-occur together in natural captions, which then leads to natural realistic prompts (\eg, \emph{a \bline{group of friends} having \bline{picnic} under a \bline{tree}}). As the \emph{realism difficulty} is increased the model generates input prompts where the co-occurence of different subjects is more and more rare, thus allowing it to generate more imaginative and challenging prompts (\eg \emph{a \bline{lion} baking \bline{cookies} in a \bline{kitchen}}). Fig.~\ref{fig:decomposable-captions-dataset-4k-realism} provides an overview of some example prompts with varying levels of realism difficulty.

\begin{figure}[t]
\begin{center}
\centerline{\includegraphics[width=1.\linewidth]{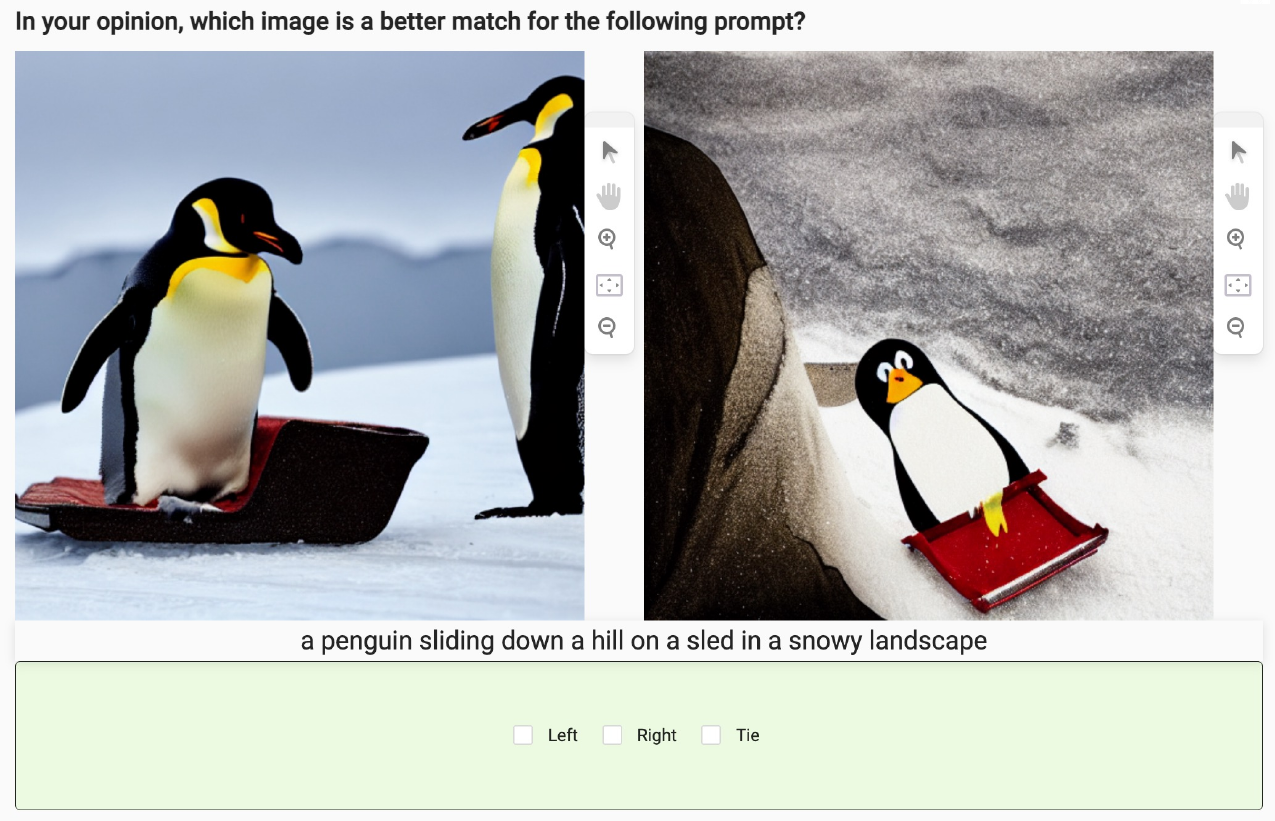}}
\caption{{\emph{Setup for pairwise user study comparing our method with prior works.}} 
}
\label{fig:pairwise-user-study}
\end{center}
\vskip -0.1in
\end{figure}

\textbf{Pairwise Human User Study.} In addition to obtaining human annotations rating (between 1 to 5) for each image-prompt pair (refer Fig.~\ref{fig:decomposable-captions-dataset-4k-rating}), we also perform a pairwise user study comparing our method with prior works.
In particular, given an input text prompt $\mathcal{P}$, the participants were shown a pair of image generation outputs comparing our method with prior works. For each pair, the human subject is then asked to select the output image which better aligns with the input prompt description. The human subjects are given three options \emph{left}, \emph{right} and \emph{tie}, where \emph{tie} indicates that both images are equally good or bad. All comparison images are generated using the same seed at $512 \times 512$ resolution with 50 inference steps for the reverse diffusion process. Fig.~\ref{fig:pairwise-user-study} provides a screenshot of the user interface for collecting the above human annotation data with pairwise comparisons.

\section{Discussion and Limitations}
\label{sec:limitations}

While the proposed iterative refinement approach shows better performance than previous works \cite{chefer2023attend,feng2022training,liu2022compositional}, it still has some limitations. 
\emph{First}, the proposed decompositional approach relies on a pretrained BLIP-VQA model \cite{li2022blip}, for determining the alignment of the generated image with each of the disjoint assertions. Thus, weaknesses of the pretrained BLIP-VQA model become our weaknesses. Recall that the VQA scores help identify the areas in which the current image generation output is lacking, which can then be addressed in the next refinement step. However, if the VQA output is not correct, then the model might focus on assertions which are already well expressed, which can reduce the efficiency of the proposed iterative refinement strategy. In future, the use of more accurate VQA models \eg, BLIP2-VQA can help alleviate this problem. Furthermore, as noted through extensive quantitative experiments across a diverse range of input prompts (refer main paper and App.~\ref{sec:decomposable-captions}), we find that the use of the BLIP-VQA model still yields quite competitive results.

\emph{Second}, without additional information from the user, the iterative refinement approach considers all assertions to be equally important in determining the overall content of the input prompt. However, as the complexity of the input prompt increases, we may wish to give more weight to certain parts of the prompt over the others. Furthermore, it is possible that the prompt may contain assertions which are not visually verifiable from the output image. For instance, for the prompt \emph{``a \bline{penguin} in a \bline{shopping mall} on a \bline{weekend}''}, the assertion about whether it is a weekend or not is not visually verifiable. Similarly certain actions \eg searching, singing are also not verifiable from a single image. In future, we would like to explore a more autonomous mechanism for including additional information like 1) user ranking for different assertions (\ie, what the user considers as important in a prompt) as well as 2) visual verifiability of a given assertion while computing the decompositional alignment scores.

\emph{Finally,} as noted in Fig.~\textcolor{red}{7} of the main paper, we note that while the proposed iterative refinement approach leads to consistent improvements in alignment accuracy over prior works, the accuracy of the alignment process decreases as the complexity of input prompt is increased. In particular, for prompts with \emph{very hard} realism difficulty, the overall alignment accuracy is only 62.9\% (Attend-and-Excite has 49.5\%). This leaves much room for improvement of text-to-image generation models. As discussed in App.~\ref{sec:comp-hps} one potential solution in this direction would be to combine recent works on human-feedback based diffusion model finetuning with the proposed training-free approach for data collection. In particular, by generating training data (on which human feedback is obtained) using the proposed iterative refinement strategy instead of previously used pretrained Stable Diffusion \cite{rombach2021highresolution} models, we can increase the quality of the finetuning process. Using the proposed decompositional alignment scores as pseudo-labels for learning the human-feedback based reward model \cite{lee2023aligning} is another interesting direction for future work. However, the same is out of scope of this paper, and we leave it as a direction for future research.

\end{document}